\documentclass[letterpaper]{article} 
\usepackage{aaai2026}  
\usepackage{times}  
\usepackage{helvet}  
\usepackage{courier}  
\usepackage[hyphens]{url}  
\usepackage{graphicx} 
\usepackage{natbib}  
\usepackage{caption} 
\usepackage{algorithm}

\usepackage{newfloat}
\usepackage{listings}
\DeclareCaptionStyle{ruled}{labelfont=normalfont,labelsep=colon,strut=off} 
\floatstyle{ruled}
\newfloat{listing}{tb}{lst}{}
\floatname{listing}{Listing}
\usepackage{amssymb}
\usepackage{graphicx}
\definecolor{mypurple}{RGB}{216,110,204} 
\definecolor{myblue}{RGB}{70,177,225}
\definecolor{mydarkgreen}{RGB}{13,161,60} 
\usepackage{colortbl}
\usepackage{multirow}
\usepackage{adjustbox} 
\usepackage[noend]{algpseudocode}
\usepackage{amsmath}
\usepackage{array}
\usepackage{multirow}
\usepackage{arydshln}
\usepackage[utf8]{inputenc}
\usepackage{csquotes} 
\usepackage{tabularx} 
\usepackage{booktabs} 
\usepackage{makecell} 
\usepackage{xcolor} 
\usepackage{tcolorbox}      
\usepackage{xurl}
\usepackage{arydshln} 
\usepackage{booktabs} 

\newtcolorbox{promptbox}{
    boxrule=1pt,                 
    arc=1mm,                       
    colback=gray!15!white,         
    colframe=gray!50!black,        
    boxsep=3pt,                    
    left=3pt, right=3pt, top=3pt, bottom=3pt, 
    fontupper=\small 
}
\definecolor{highlightcolor}{RGB}{173,216,230} 

\title{Exposing the Cracks: Vulnerabilities of Retrieval-Augmented \\LLM-based Machine Translation}
\author{
    Yanming Sun\textsuperscript{\rm 1}, Runzhe Zhan\textsuperscript{\rm 1}, Chi Seng Cheang\textsuperscript{\rm 2}, Han Wu\textsuperscript{\rm 1}, Xuebo Liu\textsuperscript{\rm 3}, Yuyao Niu\textsuperscript{\rm 4}, \\Fengying Ye\textsuperscript{\rm 1}, Kaixin Lan\textsuperscript{\rm 1}, Lidia S. Chao\textsuperscript{\rm 1}, Derek F. Wong\textsuperscript{\rm 1}\thanks{Corresponding author.}
}
\affiliations{
    \textsuperscript{\rm 1}NLP$^2$CT Lab, University of Macau\\
    \textsuperscript{\rm 2}Singapore Management University\\
    \textsuperscript{\rm 3}Harbin Institute of Technology, Shenzhen\\
    \textsuperscript{\rm 4}School of Foreign Languages, South China University of Technology\\

     nlp2ct.\{yanming, runzhe, wuhan, fengying, kaixin\}@gmail.com, cs.cheang.2025@phdcs.smu.edu.sg, \\liuxuebo@hit.edu.cn, rosenyy@scut.edu.cn, \{lidiasc, derekfw\}@um.edu.mo 
}

\usepackage{bibentry}

\begin{document}

\maketitle

\begin{abstract}
\textbf{RE}trieval-\textbf{A}ugmented \textbf{L}LM-based \textbf{M}achine \textbf{T}ranslation (REAL-MT) shows promise for knowledge-intensive tasks like idiomatic translation, but its reliability under noisy retrieval, a common challenge in real-world deployment, remains poorly understood. To address this gap, we propose a noise synthesis framework and new metrics to systematically evaluate REAL-MT’s reliability across high-, medium-, and low-resource language pairs. Using both open- and closed-sourced models, including standard LLMs and large reasoning models (LRMs), we find that models heavily rely on retrieved context, and this dependence is significantly more detrimental in low-resource language pairs, producing nonsensical translations. Although LRMs possess enhanced reasoning capabilities, they show no improvement in error correction and are even more susceptible to noise, tending to rationalize incorrect contexts. Attention analysis reveals a shift from the source idiom to noisy content, while confidence increases despite declining accuracy, indicating poor self-monitoring. To mitigate these issues, we investigate training-free and fine-tuning strategies, which improve robustness at the cost of performance in clean contexts, revealing a fundamental trade-off. Our findings highlight the limitations of current approaches, underscoring the need for self-verifying integration mechanisms.

\end{abstract}

\begin{links}
    \link{Code}{https://github.com/ymsunny/REAL-MT-Vuln}
    \link{AAAI 2026 camera-ready version}{https://aaai.org/example/camera-ready-version}
\end{links}

\begin{figure}[t]
  \centering
  \includegraphics[width=1\columnwidth]{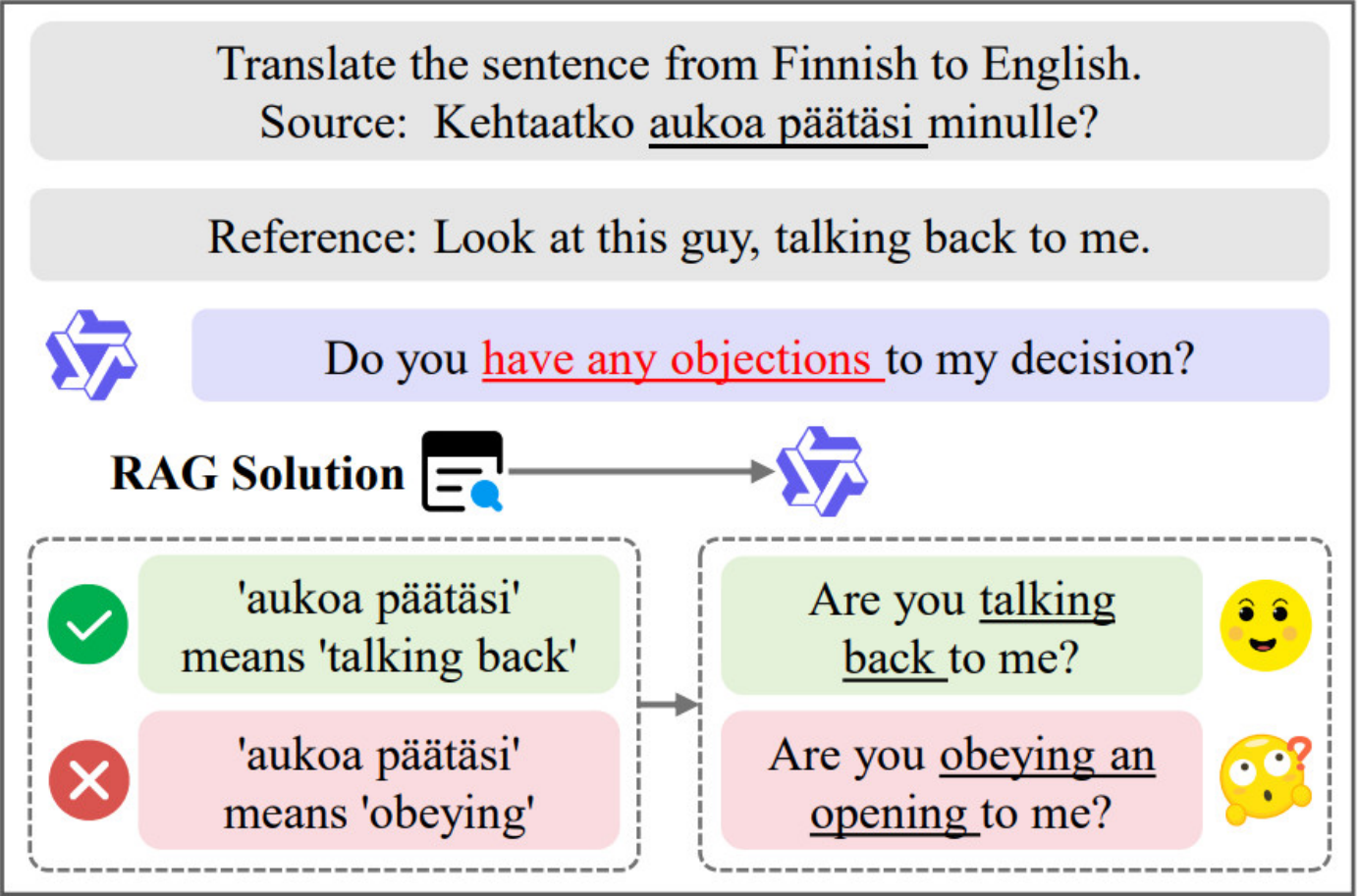}
  \caption{Examples of correct and noisy contextual cues that may arise during online retrieval. The idiom and its translation are \underline{underlined}. A robust \textbf{RE}trieval-\textbf{A}ugmented \textbf{L}LM-based \textbf{M}achine \textbf{T}ranslation (REAL-MT) system should maintain fidelity in noisy scenarios.}
  \label{fig:motivation_example}
\end{figure}

\section{Introduction}
\textbf{RE}trieval-\textbf{A}ugmented \textbf{L}LM-based \textbf{M}achine \textbf{T}ranslation (REAL-MT) is increasingly used to enhance translation quality for knowledge-intensive MT tasks like idiomatic translation~\cite{DBLP:conf/aaai/LiCYWYTX24,donthi2025improving}. Although external knowledge can enhance translation performance, reliance on it is a double-edged sword: when the retrieved context contains noise such as irrelevant or misleading information, LLMs often produce nonsensical translations, as illustrated in Figure~\ref{fig:motivation_example}. Since noisy retrieval is unavoidable in real-world deployment, a critical gap remains in understanding how REAL-MT’s reliability degrades under such conditions. This limitation constitutes a key barrier to its adoption in applications where translation quality and reliability are critical. To address this, we ask: \textbf{To what extent does noisy retrieval compromise REAL-MT’s trustworthiness?}

To systematically quantify this vulnerability, we introduce a controlled noise injection framework based on real-world retrieval failure patterns. While standard machine translation metrics like COMET~\cite{DBLP:conf/emnlp/ReiSFL20} assess overall output quality, they fail to capture semantic fidelity in idiomatic translation and cannot distinguish whether errors stem from source misinterpretation or over-reliance on retrieved context. To address this gap, we propose two complementary metrics: \textbf{Fidelity}, which evaluates translation correctness with a focus on idiomatic accuracy, and \textbf{Context Adoption Rate (CAR)}, which quantifies the extent to which models rely on external context. Together, they enable fine-grained analysis of both what the model gets wrong and why.

We evaluate REAL-MT systems instantiated with both open- and closed-source models, including standard LLMs and large reasoning models (LRM), across high-, medium-, and low-resource language pairs under controlled synthesized noise. Our results show that REAL-MT is far more vulnerable to semantic-level errors, such as irrelevant or contradictory retrieval content, than to surface-level perturbations (e.g., word reordering). Moreover, the extent of performance degradation scales with the degree of semantic deviation: the more the retrieved context diverges from the intended meaning of idioms, the more severe the drop in translation fidelity. This sensitivity is especially pronounced in low-resource language pairs, as reflected in their higher CAR, which signals greater dependence on external context and leads to more severe degradation under noisy retrieval. Surprisingly, LRM, despite enhanced reasoning capabilities, show no improvement in error correction and is even more susceptible to noise, often rationalizing incorrect contexts. 

Across noise conditions, CAR remains consistently high even when retrieved content contradicts the source, prompting us to examine whether this reflects active model integration or coincidental similarity. Our attention analysis confirms the former: models consistently attend to retrieved context regardless of its correctness, demonstrating active integration. This uncritical reliance is further exacerbated by poor metacognitive awareness, as confidence increases despite declining accuracy, a sign of \textbf{severe miscalibration and absent self-verification}.

Given this uncritical reliance on retrieved context, we explore both training-free and fine-tuning strategies to improve REAL-MT robustness. While both enhance noise resistance, they incur a consistent trade-off: performance degrades under clean contexts. Fine-tuning yields better robustness overall but still fails to adjust reliance based on context quality. This persistent trade-off reveals that the current approach cannot overcome the model’s fundamental inability to self-verify, underscoring the need for self-verifying integration mechanisms that validate retrieved content before adoption.

In summary, this work (1) first systematically evaluate REAL-MT robustness under noisy retrieval through a controlled noise injection framework, (2) proposes Fidelity and CAR as fine-grained metrics to quantify the impact of noisy retrieval on translation fidelity and context dependence, (3) reveal that REAL-MT overrelies on noisy context due to attention shifting from idioms to retrieved content, and becomes overconfident in wrong outputs, exposing poor calibration, and (4) evaluates training-free and fine-tuning mitigation strategies, revealing a fundamental trade-off between robustness under noise and performance in clean contexts, thereby pointing to the necessity of self-verifying integration mechanisms as a path forward.

\section{Related Work}
\subsection{Retrieval-Augmented LLM-based Machine Translation}
Large language models (LLMs) have revolutionized machine translation (MT), especially for low-resource languages or domains where sufficient parallel corpora are lacking~\cite{sun2024understanding,zhan2021meta,zhan2024prefix}. However, when confronted with translation scenarios demanding specific background knowledge, relying solely on the LLM's internal knowledge proves inadequate. In such instances, incorporating external knowledge to address the inherent limitations of LLMs becomes crucial~\cite{merx2024low,chen2024refining,zebaze2025compositional}. The prompt engineering capabilities of LLMs enable the integration of externally retrieved knowledge via prompting, without additional training. This makes prompt-based retrieval-augmented generation an efficient and flexible solution for MT. Recent studies by \citet{DBLP:conf/aaai/LiCYWYTX24} and \citet{donthi2024improving} leverage LLMs to enhance idiomatic translation by incorporating idiom-meaning pairs retrieved from offline knowledge bases directly into the prompt. 

Prior work operates under the assumption that all introduced external knowledge is correct, an assumption that does not hold in real-world scenarios. This work systematically investigates the impact of introducing noisy context on translation systems. Given the increasing popularity of retrieval-augmented LLM-based MT, understanding LLM performance in the face of noisy input is crucial for enhancing translation system trustworthiness.

\subsection{Robustness in Retrieval-Augmented Language Models}
Retrieval-Augmented Language Models (RALM), reliant on external retrieved content, are susceptible to compromised reliability and robustness in their generated outputs when exposed to noise, irrelevant information, or malicious data ~\citep{zhou2024trustworthiness,park2024toward,shen2024assessing,yang2025quantifying,chen2025sgic,wang2025self}. In response to the robustness challenges posed by RALMs, researchers have developed diverse approaches to enhance system robustness.~\citet{fang2024enhancing} introduces a named Retrieval-Augmented Adaptive Adversarial Training (RAAT) method to enhance the model's ability to recognize and handle various types of noise. ~\citet{yoran2023making} fine-tune the model using the parameter-efficient fine-tuning method QLoRA (Quantized Long-Range Attention) by using the synthetically generated noisy data to enhance its robustness to noisy data.~\citet{xia2025improving} introduce a novel end-to-end self-reasoning framework that enhances the robustness, interpretability, and traceability of RALMs. This improvement is achieved by leveraging the reasoning trajectories generated by the LLMs themselves.

Departing from prior research predominantly focused on English-centric scenarios, this study presents the first systematic analysis of cross-lingual translation tasks that are explicitly designed to be non-English-centric.

\section{Experimental Settings} 
\subsection{Datasets}
To analyze how resource availability affects the robustness of retrieval-augmented LLM-based machine translation (MT), we group languages into high-, medium-, and low-resource tiers following \citet{joshi2020state}, based on parallel data availability. We compile a dataset of idiomatic translations across ten language pairs, including only those data sources that provide either reference translations or explicit idiom meaning annotations,  ensuring each idiom has a grounded meaning annotation and facilitating subsequent controlled noise synthesis. The collection integrates multiple publicly available resources:

\begin{itemize}
\item \textbf{High-resource}: IdiomsInCtx-MT~\cite{stap2024fine} (English–German, German–English, Russian–English), the French–English, Japanese–English idiomatic translation from \citet{DBLP:conf/emnlp/LiuCN23}, and the KISS dataset\footnote{\url{https://github.com/Judy-Choi/KISS-Korean-english-Idioms-in-Sentences-dataSet}} for Korean–English idiomatic translation;
\item \textbf{Medium-resource}: the Finnish–English idiomatic translation from \citet{DBLP:conf/emnlp/LiuCN23};
\item \textbf{Low-resource}: the Persian–English corpus from \citet{rezaeimanesh2025large} (based on the PersianIdioms repository) and the Hindi–English corpus from \citet{donthi2025improving}.
\end{itemize}
This design enables a fine-grained analysis of REAL-MT robustness across various resource levels.

\subsection{Models}
We evaluate both open-source (Qwen series) and closed-source (claude-sonnet-4) models, covering standard LLMs (Qwen2.5-7B/14B-Instruct~\citep{qwen2.5}) and a large reasoning model (Qwen3-8B~\citep{qwen3technicalreport}), which uniquely supports seamless switching between thinking mode (enabled via \texttt{enable\_thinking=True}) and non-thinking mode (enabled via \texttt{enable\_thinking=False}). This allows us to probe whether advanced reasoning capabilities improve robustness to noisy retrieval.

Following prior work that shows lower temperatures improve translation performance~\cite{peng2023towards}, we adopt greedy decoding (\texttt{do\_sample=False}) to achieve both high output quality and reproducibility. Qwen2.5-7B-Instruct and Qwen2.5-14B-Instruct are evaluated with \texttt{max\_tokens=4096}, while Qwen3-8B uses \texttt{max\_tokens=32768} following its official configuration to use the full context length. All experiments are run with a batch size of 40 using the \texttt{vLLM}~\cite{kwon2023efficient} framework on a single NVIDIA H800 GPU with 80GB VRAM.

\subsection{Controlled Noise Context Generation}
Retrieval failures in idiomatic meaning often arise from two distinct mechanisms: incomplete phrase matching or morphological variations resulting in semantically unrelated content, while the absence of explicit idiomatic meaning online results in literal translations or low-quality paraphrases. To faithfully simulate these semantic-level errors, we design three levels of semantic noise that form a spectrum of increasing deviation from the correct idiomatic meaning:

\textbullet\ \textit{Literal Translation} ($\mathcal{N}_{\text{literal}}$): A word-by-word translation of the idiom.

\textbullet\ \textit{Semantic-Perturbed Literal Meaning} ($\mathcal{N}_{\text{semantic}}$): A variant of the literal meaning that maintains surface-level overlap but introduces a subtle semantic distortion. 

\textbullet\ \textit{Opposite Meaning} ($\mathcal{N}_{\text{opposite}}$): An adversarial variant that directly contradicts the intended meaning of the idiom.

In addition, we include syntactic perturbations (e.g., word reordering) as a control condition to isolate the impact of knowledge-level errors from surface-level input variations:

\textbullet\ \textit{Structure-Perturbed Gold Meaning} ($\mathcal{N}_{\text{struct}}$): A syntactic variant of the gold meaning, with core semantics preserved.
Together, these four noise types, summarized in Table~\ref{tab:noise_case_example}, enable a systematic investigation of how noisy retrieval compromises REAL-MT’s trustworthiness.

To balance computational cost and coverage, we randomly sample 200 instances per translation direction and use \texttt{gemini-flash-2.0} to generate noisy contexts based on carefully designed prompt templates in Appendix A.

To validate that the synthesized noisy meanings align with the intended objectives, we conduct a quantitative analysis using the following lexical and semantic metrics:

\textbullet\ \textbf{Translation Edit Rate (TER)}: Measure the degree of structural perturbation by quantifying the edit operations required to align the gold meaning ($\mathcal{G}$).

\begin{table}[t!]
\centering
\setlength{\tabcolsep}{1mm} 
\fontsize{9}{7}\selectfont
\begin{tabular}{@{}llll@{}}
\toprule
\textbf{Type} & \textbf{Text} & \textbf{Sem. Var.} & \textbf{Syn. Var.} \\
\midrule
$Idiom$           &  kankkulan kaivoon & - & -  \\
$\mathcal{G}$           & down the drain & Correct & Correct\\
$\mathcal{N}_{\text{struct}}$   & drain the down & No & Yes \\
$\mathcal{N}_{\text{literal}}$  & to Kankkula's well   & Incorrect (Relevant) & No     \\
$\mathcal{N}_{\text{semantic}}$ & to Kankkula's house  & Incorrect (Irrelevant) & No         \\
$\mathcal{N}_{\text{opposite}}$ & to good use   &  Incorrect (Opposite) & No   \\
\bottomrule
\end{tabular}
\caption{
    A case example illustrating the Finnish idiom: ``kankkulan kaivoon'', its gold meaning ($\mathcal{G}$), and four types of generated noisy meanings. ``Sem. Var.'' denotes Semantic Variations, and ``Syn. Var.'' denotes Syntactic Variations.
}
\label{tab:noise_case_example}
\end{table}

\textbullet\ \textbf{Embedding Cosine Similarity (Sim)}: Measure the semantic similarity by computing the cosine similarity between the synthesized noisy meaning and both the gold meaning ($\mathcal{G}$) and the literal translation ($\mathcal{N}_{\text{literal}}$). We use the \texttt{all-mpnet-base-v2}~\footnote{\url{https://huggingface.co/sentence-transformers/all-mpnet-base-v2}} model, which is trained on above 1 billion sentences and shows strong performance on Semantic Textual Similarity (STS) tasks. Given its effectiveness in English-centric settings, it is suitable for our evaluation, where the target language is English.

\textbullet\ \textbf{Contradiction Rate (CR)}: Measure the percentage of the generated opposite meaning ($\mathcal{N}_{\text{opposite}}$) truly contradicts the gold meaning ($\mathcal{G}$). We use the NLI model: \texttt{roberta-large-mnli}~\citep{liu2019roberta} to classify the relationship between each pair. A higher rate indicates a more effective adversarial perturbation. 

Across 10 language pairs, average TER($\mathcal{G}$, $\mathcal{N}_{\text{struct}}$) is 25.2, Sim($\mathcal{G}$, $\mathcal{N}_{\text{struct}}$) = 0.92, Sim($\mathcal{G}$, $\mathcal{N}_{\text{literal}}$) = 0.75, Sim($\mathcal{N}_{\text{literal}}$, $\mathcal{N}_{\text{semantic}}$) = 0.82, Sim($\mathcal{G}$, $\mathcal{N}_{\text{semantic}}$) = 0.73, and CR($\mathcal{G}$, $\mathcal{N}_{\text{opposite}}$) = 0.85, indicating that the generated noise aligns with the intended design. Full per-language results are presented in Table 5 in the Appendix.

\begin{table}[t]
\centering
\begin{tabular}{lcccc}
    \toprule
    \textbf{Pair} & \textbf{Metric} & \textbf{$r$} & \textbf{$\rho$} & \textbf{$\tau$} \\
    \midrule
    \multirow{3}{*}{Fi→En} & COMET-22 & 0.4179 & 0.3891 & 0.3672 \\
                                    & Fidelity & 0.9194 & 0.9286 & 0.9091 \\
    \midrule
    \multirow{3}{*}{Ja→En} & COMET-22 & 0.3191 & 0.3664 & 0.3266 \\
                                    & Fidelity & 0.8305 & 0.8196 & 0.7699  \\ 
    \midrule
    \multirow{3}{*}{Fr→En} & COMET-22 & 0.5710 & 0.5428 & 0.5270 \\
                                    & Fidelity & 0.7822 & 0.7589 & 0.7378 \\
    \bottomrule
\end{tabular}
\caption{Pearson's $r$, Spearman's $\rho$, and Kendall's $\tau$ between human evaluations and automatic metrics.}
\label{tab:metrics}
\end{table}

\begin{table*}[htbp]
\centering
{
\setlength{\tabcolsep}{1mm} 
\fontsize{9}{7}\selectfont
\begin{tabular}{c*{22}{c}}  
\toprule
\multirow{2}{*}{\textbf{Context}} & 
\multicolumn{1}{c}{\textbf{Hi}→\textbf{En}} & 
\multicolumn{2}{c}{\textbf{Fa}→\textbf{En}} & 
\multicolumn{2}{c}{\textbf{Fi}→\textbf{En}} & 
\multicolumn{2}{c}{\textbf{Ja}→\textbf{En}} & 
\multicolumn{2}{c}{\textbf{Fr}→\textbf{En}} & 
\multicolumn{2}{c}{\textbf{Ko}→\textbf{En}} & 
\multicolumn{2}{c}{\textbf{Ru}→\textbf{En}} & 
\multicolumn{2}{c}{\textbf{De}→\textbf{En}} & 
\multicolumn{2}{c}{\textbf{En}→\textbf{Fa}} & 
\multicolumn{2}{c}{\textbf{En}→\textbf{De}} & 
\multirow{2}{*}{{\textbf{Avg.\ C}}} & \multirow{2}{*}{{\textbf{ Avg.\ F}}} \\
\cmidrule(lr){2-2} \cmidrule(lr){3-4} \cmidrule(lr){5-6} \cmidrule(lr){7-8} \cmidrule(lr){9-10} \cmidrule(lr){11-12} \cmidrule(lr){13-14} \cmidrule(lr){15-16} \cmidrule(lr){17-18} \cmidrule(lr){19-20}
& \textbf{F}$\uparrow$ & \textbf{C}$\uparrow$ & \textbf{F}$\uparrow$ & \textbf{C}$\uparrow$ & \textbf{F}$\uparrow$ & \textbf{C}$\uparrow$ & \textbf{F}$\uparrow$ & \textbf{C}$\uparrow$ & \textbf{F}$\uparrow$ & \textbf{C}$\uparrow$ & \textbf{F}$\uparrow$ & \textbf{C}$\uparrow$ & \textbf{F}$\uparrow$ & \textbf{C}$\uparrow$ & \textbf{F}$\uparrow$ & \textbf{C}$\uparrow$ & \textbf{F}$\uparrow$ & \textbf{C}$\uparrow$ & \textbf{F}$\uparrow$ & & & \\
\midrule

\multicolumn{23}{c}{\textbf{\textit{Qwen2.5-7B-Instruct}}} \\
\midrule
$\mathcal{C}_\text{none}$ & \cellcolor{blue!10}0.8 & 65.8 & \cellcolor{blue!10}0.6 & 53.2 & \cellcolor{blue!5}0.4 & 59.5 & \cellcolor{blue!20}1.1 & 60.2 & \cellcolor{blue!20}1.5 & 77.6 & \cellcolor{blue!30}1.6 & 73.8 & \cellcolor{blue!30}1.8 & 71.9 & \cellcolor{blue!30}1.7 & 61.3 & \cellcolor{blue!10}0.8 & 64.5 & \cellcolor{blue!20}1.5 & 65.3 & \cellcolor{blue!20}1.2 \\
$\mathcal{G}$ & \cellcolor{blue!40}2.1 & 78.4 & \cellcolor{blue!40}2.5 & 66.0 & \cellcolor{blue!40}2.2 & 67.2 & \cellcolor{blue!40}2.4 & 67.3 & \cellcolor{blue!40}2.5 & 79.0 & \cellcolor{blue!50}2.6 & 78.9 & \cellcolor{blue!50}2.7 & 80.1 & \cellcolor{blue!50}2.7 & 63.3 & \cellcolor{blue!20}1.1 & 75.2 & \cellcolor{blue!40}2.2 & 72.8 & \cellcolor{blue!40}2.3 \\
$\mathcal{N}_{\text{struct}}$ & \cellcolor{blue!30}1.9 & 77.5 & \cellcolor{blue!40}2.2 & 63.6 & \cellcolor{blue!30}2.0 & 66.1 & \cellcolor{blue!40}2.2 & 65.4 & \cellcolor{blue!40}2.3 & 79.1 & \cellcolor{blue!40}2.5 & 77.8 & \cellcolor{blue!40}2.5 & 77.8 & \cellcolor{blue!40}2.5 & 62.6 & \cellcolor{blue!20}1.1 & 73.7 & \cellcolor{blue!40}2.2 & 71.5 & \cellcolor{blue!40}2.1 \\
$\mathcal{N}_{\text{literal}}$ & \cellcolor{blue!20}1.3 & 64.9 & \cellcolor{blue!20}1.1 & 55.4 & \cellcolor{blue!10}0.7 & 62.7 & \cellcolor{blue!20}1.5 & 58.6 & \cellcolor{blue!20}1.5 & 77.6 & \cellcolor{blue!30}1.9 & 73.7 & \cellcolor{blue!30}1.9 & 69.5 & \cellcolor{blue!30}1.6 & 59.3 & \cellcolor{blue!10}0.9 & 72.4 & \cellcolor{blue!30}2.0 & 66.0 & \cellcolor{blue!20}1.4 \\
$\mathcal{N}_{\text{semantic}}$ & \cellcolor{blue!10}0.8 & 63.3 & \cellcolor{blue!10}0.8 & 53.0 & \cellcolor{blue!5}0.5 & 60.8 & \cellcolor{blue!20}1.2 & 58.4 & \cellcolor{blue!20}1.4 & 76.5 & \cellcolor{blue!20}1.4 & 73.2 & \cellcolor{blue!30}1.6 & 67.8 & \cellcolor{blue!20}1.3 & 57.2 & \cellcolor{blue!20}1.4 & 69.1 & \cellcolor{blue!20}1.4 & 64.4 & \cellcolor{blue!20}1.2 \\
$\mathcal{N}_{\text{opposite}}$ & \cellcolor{blue!5}0.3 & 70.9 & \cellcolor{blue!5}0.5 & 57.0 & \cellcolor{blue!5}0.4 & 61.4 & \cellcolor{blue!10}0.9 & 59.7 & \cellcolor{blue!10}0.7 & 77.0 & \cellcolor{blue!20}1.4 & 73.3 & \cellcolor{blue!20}1.2 & 66.7 & \cellcolor{blue!10}0.8 & 63.3 & \cellcolor{blue!10}0.7 & 68.6 & \cellcolor{blue!20}1.2 & 66.4 & \cellcolor{blue!10}0.8 \\
\midrule

\multicolumn{23}{c}{\textbf{\textit{Qwen2.5-14B-Instruct}}} \\
\midrule
$\mathcal{C}_\text{none}$ & \cellcolor{blue!20}1.3 & 72.6 & \cellcolor{blue!20}1.2 & 57.4 & \cellcolor{blue!10}0.6 & 66.6 & \cellcolor{blue!20}1.7 & 65.1 & \cellcolor{blue!40}2.1 & 79.8 & \cellcolor{blue!30}2.0 & 77.5 & \cellcolor{blue!40}2.1 & 75.22 & \cellcolor{blue!30}1.9 & 67.2 & \cellcolor{blue!20}1.2 & 77.2 & \cellcolor{blue!30}2.0 & 71.0 & \cellcolor{blue!30}1.6 \\
$\mathcal{G}$ & \cellcolor{blue!40}2.3 & 82.2 & \cellcolor{blue!50}2.6 & 67.5 & \cellcolor{blue!40}2.4 & 68.3 & \cellcolor{blue!40}2.4 & 67.2 & \cellcolor{blue!50}2.6 & 80.1 & \cellcolor{blue!50}2.7 & 80.2 & \cellcolor{blue!50}2.7 & 80.8 & \cellcolor{blue!50}2.8 & 70.7 & \cellcolor{blue!20}1.5 & 76.8 & \cellcolor{blue!40}2.4 & 72.8 & \cellcolor{blue!40}2.3 \\
$\mathcal{N}_{\text{struct}}$ & \cellcolor{blue!40}2.1 & 80.4 & \cellcolor{blue!40}2.3 & 65.0 & \cellcolor{blue!40}2.1 & 68.2 & \cellcolor{blue!40}2.2 & 66.4 & \cellcolor{blue!40}2.4 & 80.2 & \cellcolor{blue!50}2.6 & 79.6 & \cellcolor{blue!50}2.7 & 79.8 & \cellcolor{blue!50}2.7 & 71.3 & \cellcolor{blue!20}1.5 & 77.5 & \cellcolor{blue!40}2.4 & 74.3 & \cellcolor{blue!40}2.3 \\
$\mathcal{N}_{\text{literal}}$ & \cellcolor{blue!20}1.5 & 69.0 & \cellcolor{blue!20}1.2 & 56.7 & \cellcolor{blue!10}0.7 & 64.5 & \cellcolor{blue!30}1.6 & 61.2 & \cellcolor{blue!30}1.7 & 79.1 & \cellcolor{blue!40}2.1 & 55.6 & \cellcolor{blue!40}2.1 & 72.5 & \cellcolor{blue!30}1.8 & 65.7 & \cellcolor{blue!20}1.1 & 75.1 & \cellcolor{blue!40}2.1 & 66.6 & \cellcolor{blue!30}1.6 \\
$\mathcal{N}_{\text{semantic}}$ & \cellcolor{blue!10}0.9 & 67.3 & \cellcolor{blue!10}1.0 & 54.7 & \cellcolor{blue!5}0.5 & 61.8 & \cellcolor{blue!20}1.3 & 57.3 & \cellcolor{blue!20}1.2 & 78.0 & \cellcolor{blue!30}1.7 & 73.0 & \cellcolor{blue!30}1.6 & 68.3 & \cellcolor{blue!20}1.2 & 61.1 & \cellcolor{blue!10}0.8 & 69.5 & \cellcolor{blue!20}1.5 & 65.7 & \cellcolor{blue!20}1.2 \\
$\mathcal{N}_{\text{opposite}}$ & \cellcolor{blue!5}0.5 & 74.8 & \cellcolor{blue!5}0.5 & 58.7 & \cellcolor{blue!5}0.5 & 62.2 & \cellcolor{blue!10}0.9 & 59.3 & \cellcolor{blue!10}0.6 & 78.3 & \cellcolor{blue!20}1.3 & 74.1 & \cellcolor{blue!20}1.1 & 67.5 & \cellcolor{blue!10}0.7 & 67.1 & \cellcolor{blue!10}0.7 & 71.0 & \cellcolor{blue!10}0.9 & 68.1 & \cellcolor{blue!10}0.8 \\
\midrule

\multicolumn{23}{c}{\textbf{\textit{Qwen3-8B (non-thinking mode)}}} \\
\midrule
$\mathcal{C}_\text{none}$ & \cellcolor{blue!20}1.3 & 72.7 & \cellcolor{blue!10}1.0 & 42.4 & \cellcolor{blue!10}0.6 & 64.5 & \cellcolor{blue!20}1.4 & 62.8 & \cellcolor{blue!30}1.7 & 79.6 & \cellcolor{blue!30}1.8 & 76.5 & \cellcolor{blue!30}1.8 & 73.7 & \cellcolor{blue!30}1.7 & 68.7 & \cellcolor{blue!10}1.0 & 75.2 & \cellcolor{blue!30}1.8 & 68.5 & \cellcolor{blue!20}1.4 \\
$\mathcal{G}$ & \cellcolor{blue!40}2.1 & 80.9 & \cellcolor{blue!40}2.5 & 66.0 & \cellcolor{blue!40}2.4 & 67.0 & \cellcolor{blue!40}2.3 & 66.3 & \cellcolor{blue!40}2.5 & 80.2 & \cellcolor{blue!50}2.6 & 79.3 & \cellcolor{blue!50}2.7 & 79.9 & \cellcolor{blue!50}2.7 & 71.9 & \cellcolor{blue!20}1.3 & 76.5 & \cellcolor{blue!40}2.2 & 74.2 & \cellcolor{blue!40}2.3 \\
$\mathcal{N}_{\text{struct}}$ & \cellcolor{blue!30}2.0 & 79.7 & \cellcolor{blue!40}2.2 & 62.4 & \cellcolor{blue!30}2.0 & 65.9 & \cellcolor{blue!40}2.1 & 64.4 & \cellcolor{blue!40}2.3 & 80.0 & \cellcolor{blue!40}2.5 & 78.2 & \cellcolor{blue!50}2.6 & 78.1 & \cellcolor{blue!40}2.4 & 72.1 & \cellcolor{blue!20}1.5 & 76.3 & \cellcolor{blue!40}2.2 & 73.0 & \cellcolor{blue!40}2.2 \\
$\mathcal{N}_{\text{literal}}$ & \cellcolor{blue!20}1.4 & 67.6 & \cellcolor{blue!20}1.1 & 54.9 & \cellcolor{blue!10}0.8 & 62.6 & \cellcolor{blue!20}1.5 & 56.0 & \cellcolor{blue!20}1.2 & 78.3 & \cellcolor{blue!30}2.0 & 73.9 & \cellcolor{blue!30}1.8 & 70.1 & \cellcolor{blue!30}1.6 & 68.7 & \cellcolor{blue!20}1.2 & 74.4 & \cellcolor{blue!30}2.0 & 67.4 & \cellcolor{blue!20}1.5 \\
$\mathcal{N}_{\text{semantic}}$ & \cellcolor{blue!10}0.9 & 66.2 & \cellcolor{blue!10}0.9 & 52.6 & \cellcolor{blue!5}0.5 & 59.6 & \cellcolor{blue!10}1.0 & 55.3 & \cellcolor{blue!10}1.0 & 76.5 & \cellcolor{blue!20}1.5 & 71.7 & \cellcolor{blue!20}1.3 & 65.6 & \cellcolor{blue!10}1.0 & 66.1 & \cellcolor{blue!10}1.0 & 72.2 & \cellcolor{blue!20}1.5 & 65.1 & \cellcolor{blue!20}1.1 \\
$\mathcal{N}_{\text{opposite}}$ & \cellcolor{blue!10}0.6 & 73.8 & \cellcolor{blue!5}0.5 & 56.4 & \cellcolor{blue!5}0.3 & 61.0 & \cellcolor{blue!10}0.8 & 58.0 & \cellcolor{blue!5}0.4 & 78.1 & \cellcolor{blue!20}1.5 & 72.8 & \cellcolor{blue!10}1.0 & 68.7 & \cellcolor{blue!10}0.7 & 70.4 & \cellcolor{blue!10}1.0 & 74.2 & \cellcolor{blue!20}1.5 & 68.2 & \cellcolor{blue!10}0.8 \\
\midrule

\multicolumn{23}{c}{\textbf{\textit{Qwen3-8B (thinking mode)}}} \\
\midrule
$\mathcal{C}_\text{none}$ & \cellcolor{blue!20}1.3 & 70.3 & \cellcolor{blue!10}1.0 & 55.1 & \cellcolor{blue!10}0.6 & 64.6 & \cellcolor{blue!20}1.4 & 62.6 & \cellcolor{blue!30}1.8 & 79.8 & \cellcolor{blue!30}1.9 & 75.9 & \cellcolor{blue!30}1.8 & 74.5 & \cellcolor{blue!30}1.7 & 70.7 & \cellcolor{blue!20}1.2 & 77.4 & \cellcolor{blue!30}2.0 & 70.1 & \cellcolor{blue!20}1.5 \\
$\mathcal{G}$ & \cellcolor{blue!40}2.2 & 81.9 & \cellcolor{blue!40}2.5 & 66.5 & \cellcolor{blue!40}2.4 & 66.2 & \cellcolor{blue!40}2.5 & 66.1 & \cellcolor{blue!20}1.2 & 80.3 & \cellcolor{blue!50}2.7 & 78.9 & \cellcolor{blue!50}2.7 & 79.7 & \cellcolor{blue!50}2.8 & 74.3 & \cellcolor{blue!20}1.5 & 79.8 & \cellcolor{blue!40}2.3 & 74.9 & \cellcolor{blue!40}2.3 \\
$\mathcal{N}_{\text{struct}}$ & \cellcolor{blue!30}2.0 & 80.3 & \cellcolor{blue!40}2.2 & 63.8 & \cellcolor{blue!40}2.1 & 64.7 & \cellcolor{blue!40}2.2 & 63.4 & \cellcolor{blue!40}2.3 & 79.7 & \cellcolor{blue!30}1.9 & 77.7 & \cellcolor{blue!40}2.5 & 77.4 & \cellcolor{blue!40}2.5 & 73.8 & \cellcolor{blue!20}1.5 & 79.5 & \cellcolor{blue!40}2.3 & 73.4 & \cellcolor{blue!40}2.2 \\
$\mathcal{N}_{\text{literal}}$ & \cellcolor{blue!20}1.4 & 67.2 & \cellcolor{blue!20}1.1 & 55.4 & \cellcolor{blue!10}0.8 & 61.3 & \cellcolor{blue!20}1.4 & 54.7 & \cellcolor{blue!20}1.5 & 77.5 & \cellcolor{blue!10}0.7 & 72.9 & \cellcolor{blue!30}1.8 & 68.2 & \cellcolor{blue!20}1.5 & 67.4 & \cellcolor{blue!20}1.1 & 76.2 & \cellcolor{blue!30}2.0 & 66.8 & \cellcolor{blue!20}1.3 \\
$\mathcal{N}_{\text{semantic}}$ & \cellcolor{blue!10}0.6 & 65.2 & \cellcolor{blue!10}0.8 & 53.3 & \cellcolor{blue!5}0.5 & 57.2 & \cellcolor{blue!10}0.8 & 54.0 & \cellcolor{blue!10}0.9 & 74.5 & \cellcolor{blue!20}1.2 & 70.2 & \cellcolor{blue!20}1.2 & 63.9 & \cellcolor{blue!10}1.0 & 62.7 & \cellcolor{blue!10}0.9 & 68.7 & \cellcolor{blue!20}1.2 & 63.3 & \cellcolor{blue!10}0.9 \\
$\mathcal{N}_{\text{opposite}}$ &\cellcolor{blue!5}0.1 & 73.6 & \cellcolor{blue!5}0.5 & 56.9 & \cellcolor{blue!5}0.2 & 57.9 & \cellcolor{blue!5}0.3 & 56.0 & \cellcolor{blue!5}0.1 & 76.3 & \cellcolor{blue!10}0.6 & 70.8 & \cellcolor{blue!5}0.5 & 65.0 & \cellcolor{blue!5}0.2 & 68.9 & \cellcolor{blue!5}0.5 & 70.0 & \cellcolor{blue!10}0.6 & 66.2 & \cellcolor{blue!5}0.4 \\
\bottomrule
\end{tabular}
}
\caption{Performance on idiom translation. Metrics include  Fidelity(F), Comet-22(C), with darker colors for higher scores.} 
\label{tab:multilang_evaluation}
\end{table*}

\subsection{Metrics}
\paragraph{Fidelity}
Conventional machine translation metrics like BLEU~\cite{DBLP:conf/acl/PapineniRWZ02} and COMET~\cite{DBLP:conf/emnlp/ReiSFL20} mainly measure lexical overlap or semantic alignment with reference translations. However, the key challenge in translating idioms is conveying their intended meaning. Therefore, we developed a metric specifically to \textbf{assess meaning preservation in idiom translation}, enabling more accurate evaluation of idiomatic translation quality.

Given a source sentence $x$ and retrieved context $c$, a model $M$ autoregressively generates a translation $y$. A ``fidelity'' score measuring how accurately the translation $y$ reflects the intended meaning $m$ of the source idiom.  Following previous work~\cite{DBLP:conf/emnlp/LiuIXWXZ23,DBLP:conf/aaai/LiCYWYTX24}, we use a closed-source, high-performance language model \texttt{gpt-4o-mini} for the score's evaluation. This score can be formatted as:
\begin{equation}
\mathcal{F}(y, m) = \mathop{\arg\max}\limits_{r \in \{0, 1, 2, 3\}} P(R = r \mid \text{Prompt}(y, m))
\end{equation}
where $\text{Prompt}(y, m)$ is a prompt function that generates a text prompt for the LLM, $R$ is a random variable representing the LLM's output. We generate 20 outputs per instance and take the mode as the final score. This reduces the uncertainty in the model’s scoring as an evaluator. The fidelity score ranges from 0 to 3, where 0 indicates completely unfaithful and 3 indicates perfectly faithful. Detailed prompts, variances, and standard deviations of the evaluation scores are provided in Appendix B.1 and B.3.

The REAL-MT system is robust if the value of $\mathcal{F}(y, m)$ does not decrease when the context $c$ is noisy. For instance, an LLM's robustness is demonstrated if its translation remains unaffected even when the input meaning is incorrect.

\paragraph{Context Adoption Rate (CAR)}
We design the Context Adoption Rate (CAR) metric to complement Fidelity by providing a more direct measure of a model's susceptibility to noisy context. Fidelity alone can be ambiguous, as low scores may stem from either robust rejection of noise or poor translation quality. CAR resolves this ambiguity by directly quantifying whether the model utilizes the provided context $c$ for translation, thereby enhancing the interpretability of robustness evaluations.

To calculate the CAR score, we first formalize the generation process of translation $y$ as:
\begin{equation}
P(y|x, c) = \prod_{i=1}^{n} P(y_i|y_1, y_2, ..., y_{i-1}, x, c)
\end{equation}
The generated translation $y = {y_1, y_2, ..., y_n}$, among them, each $y_i$ is the generated target word. The CAR score can be formally defined as:
\begin{equation}
\text{CAR}(c, y) = 
\begin{cases}
1, & \text{if } c \notin y_{\text{no\_context}} \land c \in y \\
0, & \text{otherwise}
\end{cases}
\end{equation}
where $y_{\text{no\_context}}$ is the translation without using context clues. We assign a score of 1 if the translation without context cues misses element $c$ while the translation using context cues includes it; otherwise, assign 0. Therefore, when $c$ is noisy, the lower the CAR score, the more robust the model is. We evaluate using \texttt{gpt-4o-mini} with prompt templates detailed in Appendix B.2.

\paragraph{Human Evaluation}
Prior work shows that reference-free metrics like CometKiwi~\citep{rei2022cometkiwi} struggle to capture idiomatic meaning~\citep{DBLP:conf/aaai/LiCYWYTX24}. To better evaluate idiom translation, we use COMET-22~\citep{rei2022comet} and introduce Fidelity as a complementary metric. We validate Fidelity via human evaluation, establishing a reliable ground truth and measuring its correlation with automatic scores. We evaluate on Fi→En, Ja→En, and Fr→En, diverse in typology and resource level. Translations are generated using Qwen2.5-7B-Instruct, Qwen2.5-14B-Instruct, and Qwen3-8B under the No Context ($\mathcal{C}_\text{none}$) setting (direct translation without external knowledge). For each language pair, the first 50 instances are annotated by three linguists following the same guidelines (see Appendix B.1) as the models. Scores are averaged for reliability. As shown in Table~\ref{tab:metrics}, Fidelity correlates highly with human judgments across languages, demonstrating that \texttt{gpt-4o-mini} is an effective automatic evaluator. We therefore adopt Fidelity as the primary metric in subsequent experiments.

To assess CAR's reliability, we conduct a human evaluation on 200 instances, achieving 82\% accuracy. Given annotation costs, we use a cost-efficient LLM-based approach (\texttt{gpt-4o-mini}). Manual analysis shows errors are mostly false 0 rather than false 1. Therefore, the high CAR values we observe under noisy contexts provide a reliable lower bound on LLM's reliance, strengthening the validity of our findings. The confusion matrix between the LLM evaluator and human annotations is provided in Appendix B.4.

\section{To What Extent Does Noisy Retrieval Compromise REAL-MT’s Trustworthiness?} \label{sec:what_extent}

We evaluate REAL-MT across six retrieval conditions: No Context ($\mathcal{C}_\text{none}$), which measures performance without external knowledge, and Gold Meaning ($\mathcal{G}$), which provides an upper bound using oracle idiomatic knowledge, along with four noise variants, Literal Translation ($\mathcal{N}_{\text{literal}}$), Semantic-Perturbed Literal Meaning ($\mathcal{N}_{\text{semantic}}$), Opposite Meaning ($\mathcal{N}_{\text{opposite}}$), and Structure-Perturbed Gold Meaning ($\mathcal{N}_{\text{struct}}$). 

\begin{figure}[t]
  \centering
  \includegraphics[width=1\columnwidth]{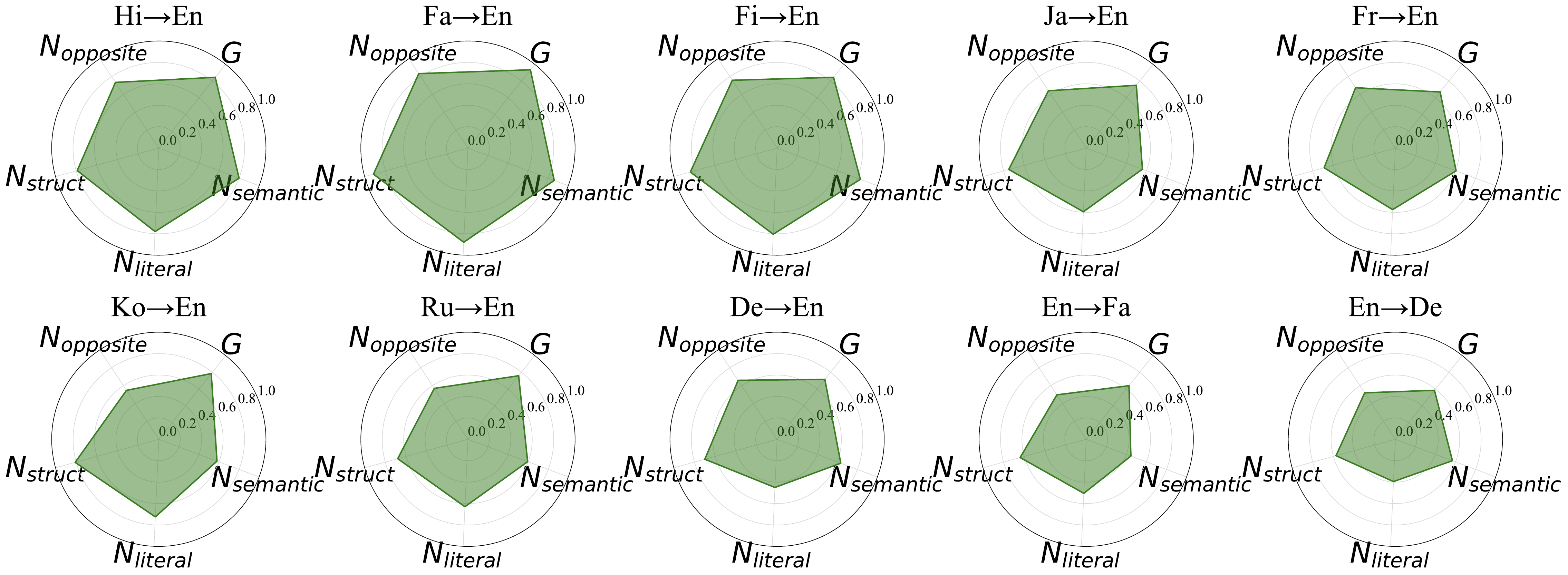}
    \caption{CAR scores for Qwen2.5-7B-Instruct under six retrieval conditions across ten language pairs.}
  \label{fig:car}
\end{figure}

\paragraph{Semantic-level noise severely undermines REAL-MT’s trustworthiness unlike surface-level perturbations.} 
As shown in Table~\ref{tab:multilang_evaluation}, models perform best with $\mathcal{G}$ context, confirming the benefit of correct context. As shown in Figure~\ref{fig:car} and Table~\ref{tab:multilang_evaluation}, both the CAR and Fidelity under $\mathcal{N}_{\text{struct}}$ is comparable to that in the $\mathcal{G}$ condition, indicating that models do not reject syntactically flawed text and possess some syntactic self-correction capabilities. With $\mathcal{N}_{\text{literal}}$, performance remains largely unchanged compared to the $\mathcal{C}_\text{none}$ setting, indicating that idiom-relevant literal meaning has limited impact on model behavior. However, when the context is irrelevant or conveys the opposite meaning, the performance significantly drops; this sensitivity increases with the degree of semantic deviation, demonstrating the model’s non-robustness to semantic noise. Similar trends are observed in our supplementary experiments on the WMT 2023 Terminology Translation dataset using Qwen2.5-7B-Instruct model, as well as on three idiomatic translation datasets evaluated with claude-sonnet-4, supporting the generalizability of our findings to other knowledge-intensive translation tasks and model architectures. The results of supplementary experiments are provided in Appendix C.

\paragraph{Large reasoning model (LRM) rationalize rather than reason in thinking mode.}
We observe that models often produce reasoning traces contradicting the noisy context, indicating awareness of inconsistency, yet generate outputs aligned with the noisy context (see (see Figure 8 in the Appendix D). This behavior suggests a tendency to \textit{rationalize} rather than \textit{reason}, potentially due to reward hacking during training, where models prioritize contextual coherence over factual fidelity~\cite{chen2025reasoning}. 
This steeper performance drop occurs in thinking mode under noisy contexts. Supporting this, results in Table~\ref{tab:multilang_evaluation} show that Qwen3-8B in non-thinking mode consistently outperforms thinking mode, underscoring the necessity of truthful inference for LRM. 


\paragraph{Low-resource languages exhibit stronger reliance on retrieval context.} 
Figure~\ref{fig:car} demonstrates that medium-to-low-resource language pairs (e.g., Hi→En, Fa→En, Fi→En) show significantly higher CAR compared to high-resource pairs (e.g., En→De, En→Fr, De→En) across both clean and noisy conditions. The pattern generalizes to other models as detailed in Appendix C. Such heightened reliance on contextual cues suggests that low-resource languages depend more heavily on external information during translation due to less parametric idiom knowledge. Consequently, LLMs become especially vulnerable to misleading information, underscoring the critical need for robustness mechanisms in low-resource settings.

\begin{figure}[t]
  \centering
  \includegraphics[width=1\columnwidth]{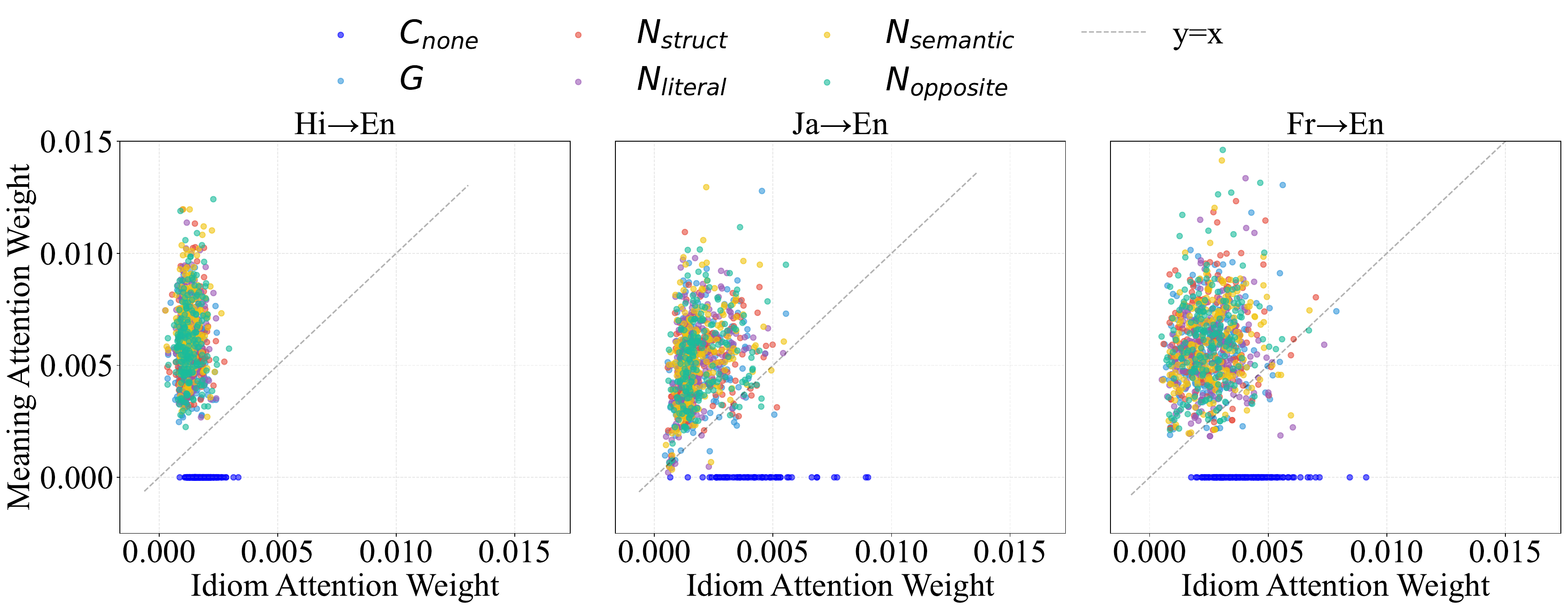}
    \caption{Attention allocation between source idiom and contextual meaning hint under various contexts.}
  \label{fig:attn_shift}
\end{figure}

\section{Uncovering the Mechanism of Context Reliance in REAL-MT}\label{sec:fail_mode}
\subsection{Attention Shift from Idiom to Retrieved Context}
LLMs exhibit high CAR under noisy contexts, indicating that their outputs are strongly influenced by retrieved content, as shown in Figure~\ref{fig:car}. However, CAR only measures output similarity and does not reveal \textit{why} the context shapes the generation process. To uncover the underlying mechanism, we analyze attention patterns during translation, as attention is widely used to diagnose models' focus during decoding~\cite{wiegreffe2019attention}. Specifically, we compute the average cumulative attention allocated to the source idiom versus the retrieved context across all target tokens.

As shown in Figure~\ref{fig:attn_shift}, attention consistently shifts toward the contextual meaning, even when it is adversarial. This systematic pattern indicates \textit{active integration}: the model consults the context during decoding, rather than merely producing aligned output. This confirms that the model's predictions are anchored in the provided context, even when they semantically contradict the source input.

\begin{table*}[t]
    \centering
    \setlength{\tabcolsep}{1mm}
    \fontsize{9}{7}\selectfont
    \begin{tabular}{c c c cc cc cc cc cc}
        \toprule
        \textbf{Language} & \textbf{Mitigation}  & {$\mathcal{C}_\text{none}$} 
        & \multicolumn{2}{c}{$\mathcal{G}$} 
        & \multicolumn{2}{c}{$\mathcal{N}_{\text{struct}}$} 
        & \multicolumn{2}{c}{$\mathcal{N}_{\text{literal}}$} 
        & \multicolumn{2}{c}{$\mathcal{N}_{\text{semantic}}$} 
        & \multicolumn{2}{c}{$\mathcal{N}_{\text{opposite}}$} \\
        \cmidrule(lr){3-3}
        \cmidrule(lr){4-5}
        \cmidrule(lr){6-7}
        \cmidrule(lr){8-9}
        \cmidrule(lr){10-11}
        \cmidrule(lr){12-13}
        \textbf{Pairs} & \textbf{Strategy}& \textbf{Fidelity} $\uparrow$ & \textbf{Fidelity} $\uparrow$ & \textbf{CAR} $\uparrow$ 
        & \textbf{Fidelity} $\uparrow$ & \textbf{CAR} $\downarrow$ 
        & \textbf{Fidelity} $\uparrow$ & \textbf{CAR} $\downarrow$ 
        & \textbf{Fidelity} $\uparrow$ & \textbf{CAR} $\downarrow$ 
        & \textbf{Fidelity} $\uparrow$ & \textbf{CAR} $\downarrow$ \\
        \midrule
        
        \multirow{5}{*}{Fr→En} 
        & Baseline & \cellcolor{blue!20}1.5 & \cellcolor{blue!40}2.5 & \cellcolor{pink!60}67\% & \cellcolor{blue!40}2.3 & \cellcolor{pink!60}69\% & \cellcolor{blue!20}1.5 & \cellcolor{pink!40}58\% & \cellcolor{blue!20}1.4 & \cellcolor{pink!60}61\% & \cellcolor{blue!10}0.7 & \cellcolor{pink!60}68\% \\
   
        & Vanilla & \cellcolor{blue!30}1.9 & \cellcolor{blue!40}2.4 & \cellcolor{pink!40}44\% & \cellcolor{blue!40}2.3 & \cellcolor{pink!40}43\% & \cellcolor{blue!20}1.5 & \cellcolor{pink!40}56\% & \cellcolor{blue!20}1.2 & \cellcolor{pink!60}65\% & \cellcolor{blue!20}1.1 & \cellcolor{pink!40}49\% \\
        & CDA     & \cellcolor{blue!30}1.9 & \cellcolor{blue!40}2.3 & \cellcolor{pink!40}32\% & \cellcolor{blue!40}2.3 & \cellcolor{pink!40}31\% & \cellcolor{blue!30}1.6 & \cellcolor{pink!40}35\% & \cellcolor{blue!30}1.6 & \cellcolor{pink!40}35\% & \cellcolor{blue!30}1.7 & \cellcolor{pink!20}21\% \\
        & ALI     & \cellcolor{blue!30}1.9 & \cellcolor{blue!40}2.2 & \cellcolor{pink!20}30\% & \cellcolor{blue!40}2.2 & \cellcolor{pink!40}31\% & \cellcolor{blue!30}1.8 & \cellcolor{pink!20}28\% & \cellcolor{blue!30}1.8 & \cellcolor{pink!20}25\% & \cellcolor{blue!30}1.8 & \cellcolor{pink!20}11\% \\
       
        & CK-PLUG & \cellcolor{blue!20}1.5 & \cellcolor{blue!40}2.4 & \cellcolor{pink!40}56\% & \cellcolor{blue!40}2.2 & \cellcolor{pink!40}53\% & \cellcolor{blue!30}1.7 & \cellcolor{pink!40}31\% & \cellcolor{blue!30}1.6 & \cellcolor{pink!40}39\% & \cellcolor{blue!20}1.3 & \cellcolor{pink!40}41\% \\
        \midrule
        
        \multirow{5}{*}{Ja→En} 
        & Baseline & \cellcolor{blue!20}1.1 & \cellcolor{blue!40}2.4 & \cellcolor{pink!60}75\% & \cellcolor{blue!40}2.2 & \cellcolor{pink!60}75\% & \cellcolor{blue!20}1.5 & \cellcolor{pink!40}60\% & \cellcolor{blue!20}1.2 & \cellcolor{pink!40}56\% & \cellcolor{blue!10}0.9 & \cellcolor{pink!60}64\% \\
    
        & Vanilla & \cellcolor{blue!20}1.4 & \cellcolor{blue!40}2.4 & \cellcolor{pink!60}69\% & \cellcolor{blue!40}2.3 & \cellcolor{pink!60}67\% & \cellcolor{blue!20}1.4 & \cellcolor{pink!60}65\% & \cellcolor{blue!20}1.1 & \cellcolor{pink!60}70\% & \cellcolor{blue!10}0.9 & \cellcolor{pink!60}62\% \\
        & CDA     & \cellcolor{blue!20}1.4 & \cellcolor{blue!30}2.0 & \cellcolor{pink!40}51\% & \cellcolor{blue!30}2.0 & \cellcolor{pink!40}54\% & \cellcolor{blue!30}1.6 & \cellcolor{pink!40} 48\% & \cellcolor{blue!20}1.3 & \cellcolor{pink!40}40\% & \cellcolor{blue!20}1.4 & \cellcolor{pink!20}23\% \\
        & ALI     & \cellcolor{blue!20}1.3 & \cellcolor{blue!30}1.9 & \cellcolor{pink!40}48\% & \cellcolor{blue!30}1.8 & \cellcolor{pink!40}47\% & \cellcolor{blue!20}1.4 & \cellcolor{pink!40}44\% & \cellcolor{blue!20}1.3 & \cellcolor{pink!40}36\% & \cellcolor{blue!20}1.4 & \cellcolor{pink!20}16\% \\
    
        & CK-PLUG & \cellcolor{blue!20}1.1 & \cellcolor{blue!40}2.1 & \cellcolor{pink!60}65\% & \cellcolor{blue!30}1.9 & \cellcolor{pink!60}70\% & \cellcolor{blue!30}1.6 & \cellcolor{pink!40}57\% & \cellcolor{blue!20}1.3 & \cellcolor{pink!40}37\% & \cellcolor{blue!20}1.3 & \cellcolor{pink!40}32\% \\
        \midrule
        
        \multirow{5}{*}{Hi→En} 
        & Baseline & \cellcolor{blue!10}0.8 & \cellcolor{blue!40}2.1 & \cellcolor{pink!60}85\% & \cellcolor{blue!30}1.9 & \cellcolor{pink!60}79\% & \cellcolor{blue!20}1.3 & \cellcolor{pink!60}78\% & \cellcolor{blue!10}0.8 & \cellcolor{pink!60}80\% & \cellcolor{blue!5}0.3 & \cellcolor{pink!60}74\% \\

        & Vanilla & \cellcolor{blue!10}0.8 & \cellcolor{blue!40}2.2 & \cellcolor{pink!60}81\% & \cellcolor{blue!30}2.0 & \cellcolor{pink!60}73\% & \cellcolor{blue!20}1.3 & \cellcolor{pink!60}72\% & \cellcolor{blue!10}0.7 & \cellcolor{pink!60}74\% & \cellcolor{blue!5}0.4 & \cellcolor{pink!60}65\% \\
        & CDA     & \cellcolor{blue!10}0.8 & \cellcolor{blue!30}2.0 & \cellcolor{pink!60}70\% & \cellcolor{blue!30}1.7 & \cellcolor{pink!60}61\% & \cellcolor{blue!20}1.2 & \cellcolor{pink!40}58\% & \cellcolor{blue!10}0.8 & \cellcolor{pink!60}62\% & \cellcolor{blue!10}0.7 & \cellcolor{pink!40}42\% \\ 
        & ALI     & \cellcolor{blue!10}0.8 & \cellcolor{blue!30}1.7 & \cellcolor{pink!40}54\% & \cellcolor{blue!20}1.5 & \cellcolor{pink!40}50\% & \cellcolor{blue!20}1.1 & \cellcolor{pink!40}52\% & \cellcolor{blue!10}0.8 & \cellcolor{pink!40}45\% & \cellcolor{blue!10}0.8 & \cellcolor{pink!20}26\% \\
       
        & CK-PLUG & \cellcolor{blue!10}0.8 & \cellcolor{blue!20}1.5 & \cellcolor{pink!60}64\% & \cellcolor{blue!30}1.6 & \cellcolor{pink!40}57\% & \cellcolor{blue!20}1.3 & \cellcolor{pink!60}64\% & \cellcolor{blue!20}1.1 & \cellcolor{pink!60}71\% & \cellcolor{blue!10}0.6 & \cellcolor{pink!40}36\% \\
        \bottomrule
    \end{tabular}
    \caption{Performance of Qwen2.5-7B-Instruct after using different mitigation strategies, evaluated on three language pairs: Fr→En (high-resource), Ja→En (medium-resource), and Hi→En (low-resource).  The darker colors indicate higher scores.}
    \label{tab:results}
\end{table*}

\subsection{Overconfidence in Context-Induced Errors}
\label{sec:confidence_change}
Given REAL-MT's uncritical adoption of noisy context, we probe the calibration of its context-induced translation. Under $\mathcal{N}_{\text{opposite}}$ noise, does the model signal awareness of its mistakes through low confidence, or does it remain highly confident in its errors, indicating poor calibration? We quantify confidence as the entropy over the output distribution, focusing on the idiom span identified via attention-based alignment. To ensure we evaluate the correct segment, we identify the target tokens corresponding to the source idiom through attention-based alignment: for each generated token, we compute its attention weights over the input and determine if it attends primarily to the idiom or contextual cue. The longest continuous sequence of idiom-aligned tokens is treated as the translated idiom span, and we report average entropy over this span.

\begin{figure}[t]
  \centering
  \includegraphics[width=1\columnwidth]{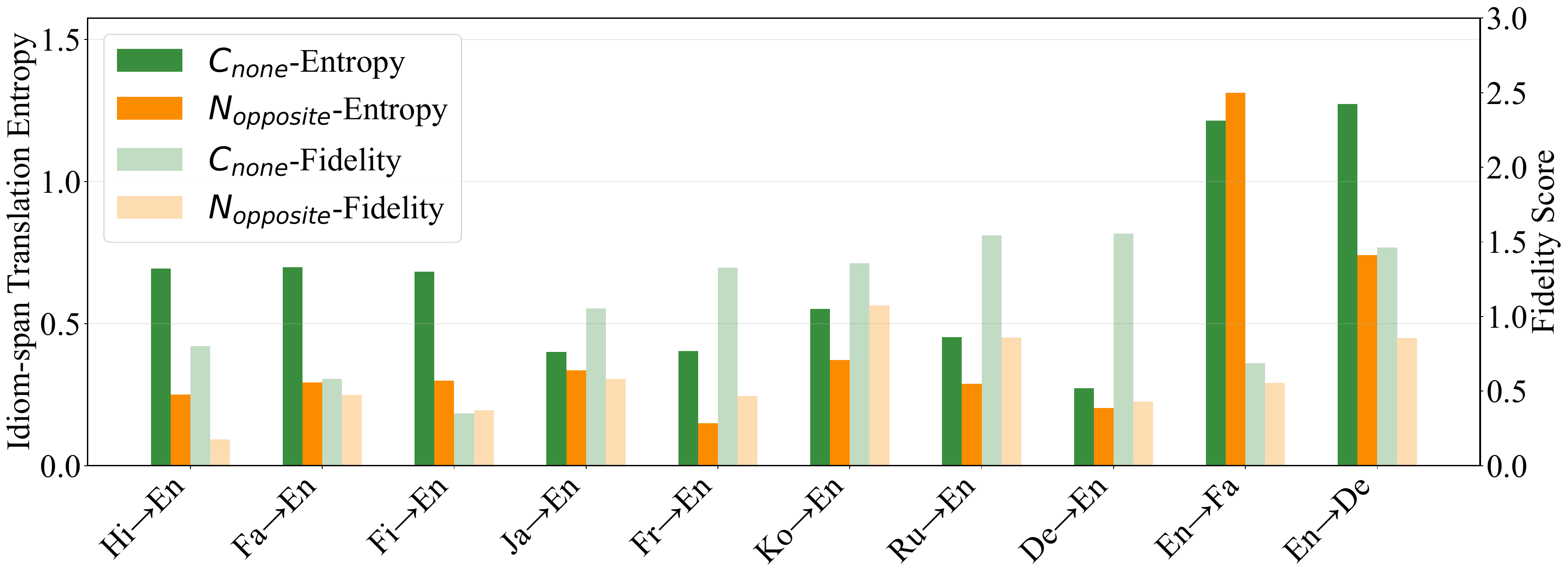}
    \caption{Comparison of confidence and Fidelity with $\mathcal{C}_\text{none}$ and $\mathcal{N}_{\text{opposite}}$ for idiom-span translations in Qwen2.5-7B-Instruct. Lower entropy implies higher confidence.}
  \label{fig:uncertainty}
\end{figure}

As shown in Figure~\ref{fig:uncertainty}, the model exhibits increased confidence in the translation of idiom spans under the $\mathcal{N}_{\text{opposite}}$ setting, even though fidelity drops significantly. This inverse relationship reveals a critical failure in calibration: the model becomes more certain of its outputs despite their inaccuracy, treating noisy context as authoritative. This highlights the model's limitations in recognizing internal and external knowledge~\cite{cheang2023can,cheang2025large}.

\section{Mitigation Strategies}
We investigate training-free and training-based methods to improve REAL-MT's robustness to noisy contexts.

\subsection{Training-free Strategy}
In RAG scenarios, prior work has explored the fusion of external knowledge to handle conflicts with internal knowledge. Given the similarity to REAL-MT, where noisy retrieved contexts may mislead translation, we investigate CK-PLUG \cite{bi2025parameters}, a training-free method that dynamically controls knowledge reliance based on context reliability. CK-PLUG computes Confidence Gain (CG) to measure the change in token-level entropy. For tokens with positive CG, which indicates the context is beneficial, the method blends the context-aware and internal distributions using $\alpha = 0.5$. For tokens with negative CG, where the context may be harmful, external knowledge is fully suppressed. We apply CK-PLUG to assess its effectiveness in improving robustness against noisy contexts in REAL-MT without requiring additional training.

\subsection{Training-based Strategy}
We investigate training-based methods to teach models to discern and reject misleading knowledge. Specifically, we construct training instances where the retrieved context is $\mathcal{N}_{\text{opposite}}$, while the target translation remains correct. This encourages the model to learn to disregard misleading external knowledge and rely more on its internal knowledge when they conflict. Our goal is to assess whether this fine-tuning scheme improves robustness to noisy contexts.
  
\subsubsection{Settings}
We perform parameter-efficient fine-tuning using Low-Rank Adaptation (LoRA)~\cite{hu2022lora}, with rank $ r = 16 $ and scaling factor $ \alpha = 16 $. Training is conducted for 50 epochs with a batch size of 2 and a learning rate of 2e-4. We use the AdamW optimizer with linear warmup and a cosine learning rate schedule. The experiments are conducted on a single NVIDIA H800 GPU with 80GB of VRAM.

\subsubsection{Dataset} We select three language pairs, Fr→En, Ja→En, and Hi→En, with varying resource levels to evaluate the generalizability of the proposed strategy across different data scales. Following~\citet{DBLP:conf/emnlp/LiuCN23}, we use their released training sets, which contain 1,000, 1,456, and 507 sentence pairs, respectively. The test sets are the same as those used in the original study.

\subsubsection{Models}
We evaluate three fine-tuning strategies, each trained on a total dataset size equal to three times the original $\mathcal{C}_\text{none}$ set to ensure fair comparison:
\begin{itemize}
\item \textbf{Vanilla}: 3$\times$ $\mathcal{C}_\text{none}$ (baseline);
\item \textbf{Adversarial Label Injection (ALI)}: 2$\times$ $\mathcal{N}_{\text{opposite}}$ + 1$\times$ $\mathcal{C}_\text{none}$;
\item \textbf{Contrastive Domain Augmentation (CDA)}: 1$\times$ $\mathcal{N}_{\text{opposite}}$ + 1$\times$ $\mathcal{G}$ + 1$\times$ $\mathcal{C}_\text{none}$.
\end{itemize}
ALI is designed to train models to discern and reject noisy texts. CDA further teaches the model discriminative reliance by enabling it to trust $\mathcal{G}$ and reject $\mathcal{N}_{\text{opposite}}$. We include $\mathcal{C}_\text{none}$ to maintain the model's core capabilities in this condition.
  
\subsection{Results and Discussion}
\paragraph{Confidence signals are unreliable for context filtering in REAL-MT.} 
As shown in Table~\ref{tab:results}, CK-PLUG yields only marginal robustness gains, revealing a fundamental flaw in entropy-based confidence signals: they assume that useful context reduces output entropy. Yet in low-resource REAL-MT, models are already overconfident in erroneous internal predictions. When accurate context ($\mathcal{G}$, e.g., Hi→En) contradicts this bias, it fails to lower entropy and may even increase uncertainty, causing CK-PLUG to suppress helpful knowledge. Rather than mitigating blind trust, this exacerbates the model’s reliance on flawed parametric knowledge, exposing the fragility of confidence-based methods in REAL-MT.

\paragraph{Training-based fine-tuning enables selective context reliance and outperforms training-free filtering.} ALI achieves the highest Fidelity and lowest CAR under $\mathcal{N}_{\text{opposite}}$
(e.g., Fr→En: Fidelity 1.8, CAR 11\%; Hi→En: Fidelity 0.8, CAR 26\%), substantially outperforming CK-PLUG. Crucially, despite being trained only on opposite-meaning noise, ALI generalizes to other semantic distortions ($\mathcal{N}_{\text{literal}}$, $\mathcal{N}_{\text{semantic}}$), demonstrating an emerging ability to discriminate reliable from misleading knowledge. However, it shows no improvement under syntactic perturbations ($\mathcal{N}_{\text{struct}}$), indicating that its robustness is limited to semantic noise and does not generalize to syntactic distortions.

\paragraph{Low-resource languages face greater difficulty in balancing noise robustness and performance with accurate retrieval.} Both training-free and training-based strategies improve robustness but degrade performance under $\mathcal{G}$, with the trade-off most pronounced in low-resource settings. This tension arises because neither approach can dynamically modulate reliance on retrieved context; instead, they apply static policies that inevitably sacrifice either noise resilience or accurate context utility. The strong dependence on hyperparameters like the blending weight $\alpha$ and training data proportions further underscores their brittleness. 

\section{Conclusions}
In this paper, we address the critical gap in understanding REAL-MT’s reliability under noisy retrieval, a common yet overlooked challenge in real-world deployment. To this end, we propose a controlled noise synthesis framework and two new metrics, Fidelity and Context Adoption Rate (CAR), to systematically evaluate REAL-MT’s robustness in knowledge-intensive translation tasks. Our evaluation, using both open- and closed-sourced models across high-, medium-, and low-resource language pairs, reveals that all models exhibit a high sensitivity to semantic noise. This is especially pronounced in low-resource scenarios, where models are more likely to uncritically rely on the noisy external context, often leading to nonsensical translations. Our experiments on LRM uncover that it is more vulnerable to noisy context than the non-reasoning version and, critically, tend to rationalize the noise even when aware of it, failing to demonstrate truthful reasoning. An analysis of attention mechanisms and model uncertainty reveals two failures: a shift of attention away from source idioms towards noisy context, compounded by a rise in model confidence despite a decline in accuracy. This points to critically weak self-verification and poor calibration. While both training-free and fine-tuning strategies can enhance robustness under noisy contexts, they do so at the cost of performance in accurate contexts. This trade-off underscores the necessity of developing models with self-reflective capabilities.

\section{Acknowledgments}
This work was supported in part by the Science and Technology Development Fund of Macau SAR (Grant Nos. FDCT/0007/2024/AKP, FDCT/0070/2022/AMJ, FDCT/060/2022/AFJ), the National Natural Science Foundation of China (Grant Nos. 62261160648, 62266013), the China Strategic Scientific and Technological Innovation Cooperation Project (Grant No. 2022YFE0204900), and the UM and UMDF (Grant Nos. MYRG-GRG2025-00236-FST, MYRG-GRG2023-00006-FST-UMDF, MYRG-GRG2024-00165-FST-UMDF, EF2024-00185-FST, EF2023-00151-FST).



\section*{A Prompt Templates of Controlled Noise Context Generation}
\label{app:noise_generation}

\subsection*{A.1 Gold Meaning Extraction}
\begin{promptbox}
Provide the meaning of the idiom ``\{idiom\_in\_source\}'' as used in the source sentence, based on the context of the sentence and its translation.

Source Sentence: \{source\}

Translation: \{target\}

Respond only with the idiom's meaning in English without any explanation, formatted as a JSON object: 

\{ \\
``output'': ``meaning content'' \\ 
\}
\end{promptbox}

\subsection*{A.2 Structure-Perturbed Gold Meaning Generation}
\begin{promptbox}
Goal: Create text with SUBTLE GRAMMATICAL ERRORS while preserving the core meaning of the input.

Input: \{groundtruth\_meaning\}

Methods (MUST apply at least one):

1. Swap adjacent phrases/clauses to create awkward phrasing (e.g., ``New problems that will arise'' → ``Problems new that will arise'')

2. Move modifier to incorrect position (e.g., ``The quickly running child'' → ``The running quickly child'')

3. Slight word order inversion in short phrases (e.g., ``good ideas'' → ``ideas good'')

Constraints:

1. CRITICAL: Output MUST contain subtle grammatical errors

2. The output MUST be different from the input, even for very short phrases

3. Do NOT change the vocabulary - use the same words but in incorrect order/structure

4. For extremely short inputs (2-3 words), apply method \#3 aggressively

**Important:** Provide your response in the following strict JSON format:

\{\\
``output'': ``Your output with subtle grammatical errors''\\
\}
\end{promptbox}

\subsection*{A.3 Literal Meaning Generation}
\begin{promptbox}
Given the idiom or phrase ``\{idiom\_in\_source\}'', please do the following:

1. Provide a literal translation of the idiom (word-for-word translation)

2. Generate a phrase with similar meaning to the literal translation (NOT the idiomatic meaning)

For example:

- Literal translation: ``to hit the grass and startle the snake''
- Similar phrase: ``to disturb the vegetation and alarm the reptile''

- Literal translation: ``to draw a snake and add feet''
- Similar phrase: ``to sketch a serpent and attach limbs''

Please respond in English and format it as the following JSON object:

\{\\
``literal\_translation'': ``the word-for-word translation'',\\
``similar\_phrase'': ``phrase with similar meaning to the literal translation''\\
\}
\end{promptbox}

\begin{table*}[t]
\centering
\setlength{\tabcolsep}{1mm} 
\fontsize{9}{9}\selectfont
\begin{tabular}{lcccccc}
\toprule
\textbf{Language Pair} & 
\textbf{TER}($\mathcal{G}, \mathcal{N}_{\text{struct}}$) $\downarrow$ & 
\textbf{Sim}($\mathcal{G}, \mathcal{N}_{\text{struct}}$) $\uparrow$ & 
\textbf{Sim}($\mathcal{G}$, $\mathcal{N}_{\text{literal}}$) &
\textbf{Sim}($\mathcal{N}_{\text{literal}}$, $\mathcal{N}_{\text{semantic}}$) $\uparrow$ &
\textbf{Sim}($\mathcal{G}$, $\mathcal{N}_{\text{semantic}}$) $\downarrow$ &
\textbf{CR}($\mathcal{G}, \mathcal{N}_{\text{opposite}}$) $\uparrow$  \\
\midrule
Hi $\rightarrow$ En & 28.6 & 0.91 & 0.75 & 0.82 & 0.68 & 0.65 \\
Fa $\rightarrow$ En & 26.1 & 0.91 & 0.68 & 0.79 & 0.67 & 0.74 \\
Fi $\rightarrow$ En & 27.5 & 0.87 & 0.55 & 0.71 & 0.52 & 0.63 \\
Ja $\rightarrow$ En & 33.7 & 0.88 & 0.70 & 0.79 & 0.64 & 0.68 \\
Fr $\rightarrow$ En & 20.5 & 0.91 & 0.66 & 0.77 & 0.66 & 0.70 \\
Ko $\rightarrow$ En & 21.0 & 0.95 & 0.89 & 0.91 & 0.87 & 0.87 \\
Ru $\rightarrow$ En & 12.9 & 0.96 & 0.85 & 0.90 & 0.83 & 0.87 \\
De $\rightarrow$ En & 18.6 & 0.92 & 0.70 & 0.79 & 0.68 & 0.72 \\
En $\rightarrow$ Fa & 43.6 & 0.95 & 0.92 & 0.92 & 0.91 & 0.92 \\
En $\rightarrow$ De & 19.3 & 0.92 & 0.84 & 0.82 & 0.79 & 0.80 \\
\midrule
\textbf{Avg.} & \textbf{25.2} & \textbf{0.92} & \textbf{0.75} & \textbf{0.82} & \textbf{0.73} & \textbf{0.85}\\
\bottomrule
\end{tabular}
\caption{
Quantitative evaluation of perturbed meanings across 10 language pairs. We report Translation Edit Rate (TER), embedding-based similarity (Sim), and Contradiction Rate (CR) for key comparisons.
}
\label{tab:noise_evaluation}
\end{table*}

\subsection*{A.4 Semantic-Perturbed Literal Meaning Generation}
\begin{promptbox}
I have a literal translation of an idiom: "{literal\_meaning}"

Please replace ONE key noun in this phrase with a different noun to significantly change its meaning.

For example: \\
- "to hit the grass and startle the snake" → "to hit the tree and startle the snake" \\
- "to draw a snake and add feet" → "to draw a snake and add wings" \\
- "to bite the bullet" → "to bite the apple" \\
- "hit your head" → "hit your face" \\

Guidelines: \\
1. Replace ONLY ONE noun (person, place, thing, or concept) \\
2. The substitution should significantly change the meaning \\
3. Keep the grammatical structure intact \\
4. Make sure the result is still grammatically correct \\

Please format your response as a JSON object:
\{\\
    ``output'': ``The modified literal translation with exactly one noun replaced'' \\
\}
\end{promptbox}

\subsection*{A.5 Opposite Meaning Generation}
\begin{promptbox}
Provide an opposite meaning for: \{groundtruth\_meaning\}

Respond only with the opposite meaning in English without any explanation, formatted as a JSON object:

\{\\``output'': ``opposite meaning''\\\}
\end{promptbox}

\section*{B Metrics}
\label{app:metrics}
We use open-source LLMs with prompting for automated evaluation.

\subsection*{B.1 Fidelity}
\label{app:evaluation_guidelines}

\begin{promptbox}
Please evaluate the translation quality of the machine translation in conveying the given meaning.

Machine translation: \{translation\}

Meaning: \{meaning\}

Evaluation criteria:

Based on the machine translation, does it accurately convey the following meaning: \{meaning\}?

Please rate the following dimension (0-3 points) and provide a detailed explanation:

Fidelity: To what extent does the machine translation accurately convey the meaning provided?

0 points: Conveys a meaning that is opposite or contradictory to the provided meaning.

1 point: Does not convey any aspect of the provided meaning; the translation is non-idiomatic or literal and misses the intended sense.

2 points: Partially conveys the meaning.

3 points: Perfectly conveys the meaning.

Please provide your response in this JSON format:

\{\\
``Fidelity'': \{\\
``score'': Rating from 0-3,\\
``justification'': ``[Brief justification for your rating]''\\\}\\
\} 
\end{promptbox}

\subsection*{B.2 Context Adoption Rate}

\begin{promptbox}
You will receive a meaning (Provided Meaning), an original translation (Original Translation), and a new translation (Translation). Your task is to evaluate whether the Translation effectively uses the Provided Meaning, specifically by incorporating information or nuances present in the Provided Meaning that are *not* present in the Original Translation.

Provided Meaning: \{meaning\} \\
Original Translation: \{original\_translation\} \\
Translation: \{translation\} \\

Evaluation Criteria:

If the Translation incorporates elements of the Provided Meaning that are absent in the Original Translation, give a score of 1, else 0.

Please respond using the following JSON format: \\
\{\\
``score'': 0 or 1, \\
``explanation'': ``Analysis of how the Translation incorporates elements of the Provided Meaning, especially highlighting elements present in the Meaning but absent in the Original Translation.  Explain why the score was assigned.'', \\
``evidence'': ``The portion of the Translation that uses the elements from the meaning, and a comparison to the corresponding (or absent) portion in the Original Translation.'' \\
\} 
\end{promptbox}

\subsection*{B.3 Variance and Standard Deviation of Fidelity Metrics in LLM Evaluations}
We examine the variance and standard deviation of fidelity metrics derived from evaluations conducted by the large language model, with the temperature set to 1. The statistical results are based on all participating samples in the evaluation. As shown in Table~\ref{tab:variance}, these statistical measures provide insights into the consistency and reliability of LLM-as-a-judge.

\begin{table}[t]
\centering
\begin{tabular}{cc|cc}
\hline
\multicolumn{2}{c|}{} & \multicolumn{2}{c}{\textbf{LLM}} \\
\multicolumn{2}{c|}{} & \textbf{CAR=0} & \textbf{CAR=1} \\
\hline
\multirow{2}{*}{\textbf{Human}} 
& \textbf{CAR=0} & 18 & 6 \\
& \textbf{CAR=1} & 31 & 145 \\
\hline
\end{tabular}
\caption{Confusion Matrix of CAR Evaluation}
\label{tab:confusion_matrix}
\end{table}

\begin{table}[t]
\centering
\setlength{\tabcolsep}{1mm} 
\fontsize{9}{7}\selectfont
\begin{tabular}{l*{6}{c}}
\toprule
\multirow{2}{*}{\textbf{Context}} & \multicolumn{2}{c}{\textbf{Zh}→\textbf{En}} & 
\multicolumn{2}{c}{\textbf{De}→\textbf{En}} & 
\multicolumn{2}{c}{\textbf{En}→\textbf{Cs}} \\
\cmidrule(lr){2-3} \cmidrule(lr){4-5} \cmidrule(lr){6-7}
& \textbf{Fidelity}$\uparrow$ & \textbf{CAR} & \textbf{Fidelity}$\uparrow$ & \textbf{CAR} & \textbf{Fidelity}$\uparrow$  & \textbf{CAR}  \\
\midrule
$\mathcal{C}_\text{none}$ & \cellcolor{blue!20}1.5 & - &  \cellcolor{blue!40}2.2 & - &  \cellcolor{blue!40}2.3 & -   \\
$\mathcal{G}$ & \cellcolor{blue!50}2.6 & \cellcolor{pink!60}80\% & \cellcolor{blue!50}2.8 & \cellcolor{pink!40}57\% &  \cellcolor{blue!50}2.9 & \cellcolor{pink!60}77\% \\
$\mathcal{N}_{\text{struct}}$ & \cellcolor{blue!30}2.0  & \cellcolor{pink!60}86\% & \cellcolor{blue!40}2.5 & \cellcolor{pink!40}60\% &  \cellcolor{blue!50}2.6 & \cellcolor{pink!60}88\% \\
$\mathcal{N}_{\text{literal}}$ & \cellcolor{blue!20}1.5 & \cellcolor{pink!60}76\% & \cellcolor{blue!40}2.1 & \cellcolor{pink!40}43\% &  \cellcolor{blue!40}2.4 & \cellcolor{pink!60}82\% \\
$\mathcal{N}_{\text{semantic}}$ & \cellcolor{blue!20}1.2  & \cellcolor{pink!60}89\% & \cellcolor{blue!30}1.8 & \cellcolor{pink!60}88\% &  \cellcolor{blue!30}1.8 & \cellcolor{pink!100}91\% \\
$\mathcal{N}_{\text{opposite}}$ & \cellcolor{blue!10}0.7 & \cellcolor{pink!60}84\% & \cellcolor{blue!20}1.3 & \cellcolor{pink!60}75\% &  \cellcolor{blue!20}1.2 & \cellcolor{pink!100}98\% \\
\bottomrule
\end{tabular}
\caption{Evaluation Results from Qwen2.5-7B-Instruct.}
\label{tab:terminology_evaluation}
\end{table}

\begin{table}[t]
\centering
\setlength{\tabcolsep}{1mm} 
\fontsize{9}{7}\selectfont
\begin{tabular}{l*{6}{c}}
\toprule
\multirow{2}{*}{\textbf{Context}} & \multicolumn{2}{c}{\textbf{Hi}→\textbf{En}} & 
\multicolumn{2}{c}{\textbf{Ja}→\textbf{En}} & 
\multicolumn{2}{c}{\textbf{Fr}→\textbf{En}} \\
\cmidrule(lr){2-3} \cmidrule(lr){4-5} \cmidrule(lr){6-7}
& \textbf{Fidelity}$\uparrow$ & \textbf{CAR} & \textbf{Fidelity}$\uparrow$ & \textbf{CAR} & \textbf{Fidelity}$\uparrow$  & \textbf{CAR}  \\
\midrule
\multicolumn{7}{c}{\textbf{\textit{DeepSeek-R1-Distill-Llama-8B}}} \\
\midrule
$\mathcal{C}_\text{none}$ & \cellcolor{blue!10}0.6 & - &  \cellcolor{blue!10}1.0 & -  &  \cellcolor{blue!20}1.2 & -   \\
$\mathcal{G}$ & \cellcolor{blue!30}2.0 & \cellcolor{pink!100}93\% & \cellcolor{blue!40}2.2  & \cellcolor{pink!100}93\% &  \cellcolor{blue!40}2.4 & \cellcolor{pink!60}86\% \\
$\mathcal{N}_{\text{struct}}$ & \cellcolor{blue!30}1.9  & \cellcolor{pink!100}92\% & \cellcolor{blue!30}1.9 & \cellcolor{pink!100}95\% &  \cellcolor{blue!40}2.1 & \cellcolor{pink!60}87\% \\
$\mathcal{N}_{\text{literal}}$ & \cellcolor{blue!10}0.9 & \cellcolor{pink!100}95\% & \cellcolor{blue!20}1.3 & \cellcolor{pink!100}92\% &  \cellcolor{blue!20}1.2 & \cellcolor{pink!60}73\% \\
$\mathcal{N}_{\text{semantic}}$ & \cellcolor{blue!10}0.7  & \cellcolor{pink!100}95\% & \cellcolor{blue!20}1.3 & \cellcolor{pink!100}96\% &  \cellcolor{blue!20}1.1 & \cellcolor{pink!60}89\% \\
$\mathcal{N}_{\text{opposite}}$ & \cellcolor{blue!5}0.4 & \cellcolor{pink!60}90\% & \cellcolor{blue!10}0.6 & \cellcolor{pink!60}85\% &  \cellcolor{blue!10}0.8 & \cellcolor{pink!60}76\% \\
\midrule
\multicolumn{7}{c}{\textbf{\textit{Claude}}} \\
\midrule
$\mathcal{C}_\text{none}$ & \cellcolor{blue!30}1.6 & - &  \cellcolor{blue!40}2.2 & - &  \cellcolor{blue!40}2.4 & -   \\
$\mathcal{G}$ & \cellcolor{blue!50}2.6 & \cellcolor{pink!60}75\% & \cellcolor{blue!40}2.1 & \cellcolor{pink!60}69\% &  \cellcolor{blue!50}2.7 & \cellcolor{pink!40}51\% \\
$\mathcal{N}_{\text{struct}}$ & \cellcolor{blue!40}2.5  & \cellcolor{pink!60}73\% & \cellcolor{blue!40}2.4 & \cellcolor{pink!60}65\% &  \cellcolor{blue!50}2.6 & \cellcolor{pink!40}52\% \\
$\mathcal{N}_{\text{literal}}$ & \cellcolor{blue!20}1.2 & \cellcolor{pink!60}62\% & \cellcolor{blue!30}1.9 & \cellcolor{pink!40}55\% &  \cellcolor{blue!40}2.1 & \cellcolor{pink!40}31\% \\
$\mathcal{N}_{\text{semantic}}$ & \cellcolor{blue!20}1.2  & \cellcolor{pink!60}70\% & \cellcolor{blue!30}1.9 & \cellcolor{pink!60}61\% &  \cellcolor{blue!30}1.9 & \cellcolor{pink!40}39\% \\
$\mathcal{N}_{\text{opposite}}$ & \cellcolor{blue!10}0.8 & \cellcolor{pink!60}88\% & \cellcolor{blue!20}1.2 & \cellcolor{pink!60}65\% &  \cellcolor{blue!20}1.2 & \cellcolor{pink!60}68\% \\
\bottomrule
\end{tabular}
\caption{Evaluation Results from DeepSeek-R1-Distill-Llama-8B and Claude.}
\label{tab:other_models}
\end{table}

\begin{table*}[t]
\centering
{
\setlength{\tabcolsep}{1mm} 
\fontsize{9}{7}\selectfont
\begin{tabular}{c*{23}{c}}  
\toprule
\multirow{2}{*}{\textbf{Context}} & 
\multicolumn{2}{c}{\textbf{Hi}→\textbf{En}} & 
\multicolumn{2}{c}{\textbf{Fa}→\textbf{En}} & 
\multicolumn{2}{c}{\textbf{Fi}→\textbf{En}} & 
\multicolumn{2}{c}{\textbf{Ja}→\textbf{En}} & 
\multicolumn{2}{c}{\textbf{Fr}→\textbf{En}} & 
\multicolumn{2}{c}{\textbf{Ko}→\textbf{En}} & 
\multicolumn{2}{c}{\textbf{Ru}→\textbf{En}} & 
\multicolumn{2}{c}{\textbf{De}→\textbf{En}} & 
\multicolumn{2}{c}{\textbf{En}→\textbf{Fa}} & 
\multicolumn{2}{c}{\textbf{En}→\textbf{De}} \\
\cmidrule(lr){2-3} \cmidrule(lr){4-5} \cmidrule(lr){6-7} \cmidrule(lr){8-9} \cmidrule(lr){10-11} \cmidrule(lr){12-13} \cmidrule(lr){14-15} \cmidrule(lr){16-17} \cmidrule(lr){18-19} \cmidrule(lr){20-21}
& \textbf{Var} & \textbf{SD} & \textbf{Var} & \textbf{SD} & \textbf{Var} & \textbf{SD} & \textbf{Var} & \textbf{SD} & \textbf{Var} & \textbf{SD} & \textbf{Var} & \textbf{SD} & \textbf{Var} & \textbf{SD} & \textbf{Var} & \textbf{SD} & \textbf{Var} & \textbf{SD} & \textbf{Var} & \textbf{SD}\\
\midrule

\multicolumn{22}{c}{\textbf{\textit{Qwen2.5-7B-Instruct}}} \\
\midrule
$\mathcal{C}_\text{none}$ & 0.05 & 0.13 & 0.05 & 0.12 & 0.06  & 0.15  & 0.10  & 0.23  & 0.06 & 0.15 & 0.05 &0.14 &	0.05 &	0.13 &0.06 &	0.14 & 0.06 &0.16 &	0.06 & 0.15\\
$\mathcal{G}$ & 0.06 & 0.14 & 0.06 & 0.14 & 0.07 & 0.15 & 0.05 & 0.11 & 0.04 
 & 0.09 & 0.04 &0.11 &0.04 &	0.09 &	0.04 &	0.10 &0.07 &0.16 &0.06 & 0.15\\
$\mathcal{N}_{\text{struct}}$ & 0.05 & 0.13 & 0.05 & 0.13 & 0.06 & 0.14 & 0.05 & 0.12 & 0.05 & 0.11 & 0.05 & 0.13 & 0.04 & 0.10 & 0.04 & 0.11&  0.07 & 0.17 &0.07 & 0.17\\
$\mathcal{N}_{\text{literal}}$ & 0.05 & 0.12 & 0.05 & 0.13 & 0.06  & 0.15 & 0.04 
 & 0.10  & 0.05 & 0.12 & 0.08 &0.19 & 0.10 & 0.13 &	0.04 & 0.12 & 0.17	& 0.15 & 0.07 & 0.17\\
$\mathcal{C}_{\text{semantic}}$ & 0.05 & 0.12 & 0.06 & 0.14 & 0.07 & 0.15  & 0.05 & 0.13 & 0.04  & 0.10  & 0.06 &0.16 &0.06 &0.14 &0.04 &0.11 &0.07 &0.16&0.06 & 0.16\\
$\mathcal{N}_{\text{opposite}}$ & 0.04 & 0.09 & 0.05 & 0.11 & 0.04 & 0.09  & 0.04 & 0.11 & 0.04 & 0.10 & 0.07 &0.17 &0.06 &0.13 &0.04  &0.10 &0.05  &0.12 &0.05 & 0.13\\
\midrule

\multicolumn{22}{c}{\textbf{\textit{Qwen2.5-14B-Instruct}}} \\
\midrule
$\mathcal{C}_\text{none}$ & 0.06 &0.15 &0.05 &0.13 &0.06&0.14 &0.04 &0.10 &0.05 &0.12 &0.06 &0.15 & 0.05 
&0.12 &	0.05 &0.13 &	0.06 &0.16 &0.07 &0.17 \\
$\mathcal{G}$ &0.06 &0.14 &	0.06 &0.14 	&0.05 &	0.12 &0.04 	&0.09 &0.04 &0.09 &0.05 &0.11 &0.04 
& 0.09 &0.03 &0.07 &0.07 &0.19 	&0.07 &0.17 \\
$\mathcal{N}_{\text{struct}}$ & 0.06 &0.14 &0.05 &0.13 &0.06 &0.14 &0.04 &0.10 &0.04 &0.11 &0.05 &0.12 & 0.04 
&0.11 &0.03 &	0.08 &0.07 &0.18 &0.07 &0.17 \\
$\mathcal{N}_{\text{literal}}$ & 0.06 &0.14 &0.07 &0.17 &0.06 &0.13 &0.05 &	0.12 &0.05 &0.12 &0.07 &0.16 &0.05 
&0.12  &0.05 &0.12 &0.07 &0.17 &0.07 &0.18 \\
$\mathcal{C}_{\text{semantic}}$ & 0.06 &0.14 &0.06 	&0.15 &0.06 &0.06 &0.05 &0.13 &0.05 &0.14 &0.06 &0.16 &0.06  &0.15 
&0.04 & 0.10 &0.06 &0.14 &0.07 &0.17 \\
$\mathcal{N}_{\text{opposite}}$ & 0.03 &0.09 &0.06 &0.13 &0.03 &0.08 &0.05 	&0.11 &0.04 &0.09 &0.07 &0.12 &0.05 &0.12  &0.04 &0.10 &0.06 &0.15 &0.05 &0.12 \\
\midrule

\multicolumn{22}{c}{\textbf{\textit{Qwen3-8B (non-thinking mode)}}} \\
\midrule
$\mathcal{C}_\text{none}$ & 0.07 & 0.17 & 0.04 & 0.11 & 0.07 & 0.16 & 0.04 
 & 0.11 & 0.05 & 0.13 & 0.07& 0.18 &0.05 &0.12 &0.05 &0.12 &0.07 &0.17 &0.07 &0.17 
 \\
$\mathcal{G}$ & 0.06 & 0.16 & 0.05 & 0.13 & 0.06 & 0.13 & 0.04 & 0.09 & 0.04  & 0.09 & 0.05 &0.12 &0.04 &0.10 &0.04 &	0.09 &0.06 &0.15 &0.06 &0.16 \\
$\mathcal{N}_{\text{struct}}$ & 0.06 & 0.14 & 0.05 & 0.13 &0.07 &0.16 & 0.04
& 0.10 & 0.05 & 0.10 & 0.07 &0.17 	&0.06 &0.13 &0.04 &0.11 &0.08 &0.20 &0.07 &0.18 \\
$\mathcal{N}_{\text{literal}}$ & 0.05 & 0.12 & 0.05 & 0.13 & 0.06 & 0.14 & 0.05 
& 0.13 & 0.05 & 0.13 & 0.06 &0.15& 0.05&	0.14 &0.04 &0.10 &0.06 &0.16&0.06 &0.15 \\
$\mathcal{C}_{\text{semantic}}$ & 0.05 & 0.13 & 0.05 & 0.14 & 0.06 & 0.14 & 0.05 
& 0.13 & 0.05 & 0.14  & 0.06 &0.15 &0.05 &0.13 &0.05 &0.12 &0.06 &0.15 	& 0.05 &0.14 \\
$\mathcal{N}_{\text{opposite}}$ & 0.04 & 0.09 & 0.04 & 0.11 & 0.04 & 0.10 & 0.04  & 0.09 & 0.04 & 0.09 & 0.07 &0.17 &0.05 &0.12 &0.04 &0.11 	&0.05 &0.12 &0.07 &0.16\\
\midrule

\multicolumn{22}{c}{\textbf{\textit{Qwen3-8B (thinking mode)}}} \\
\midrule
$\mathcal{C}_\text{none}$ & 0.05 & 0.13 & 0.05 & 0.13 & 0.06 & 0.14 & 0.05 
 & 0.13 & 0.05 & 0.14 & 0.06 &0.16 	&0.05 &0.13 &0.05 &0.12 &0.07 	&0.18 	&0.07 &0.17 \\
$\mathcal{G}$ & 0.07 & 0.16 & 0.04 & 0.10 &0.07 & 0.16 & 0.04 & 0.10 & 0.05 
& 0.11 & 0.05 &0.13 &0.05 &0.10 &0.03 &0.09 &0.07 &0.18 &0.07 &0.16 \\
$\mathcal{N}_{\text{struct}}$ & 0.06 & 0.15 &0.05 & 0.12 & 0.07 & 0.16 & 0.05 & 0.12 & 0.05 & 0.12 & 0.06 &0.14 &0.06 &0.14 &0.04 &0.09 &0.07 &0.18 &0.07 &0.17 \\
$\mathcal{N}_{\text{literal}}$ & 0.04 &0.11 &0.05 & 0.12 &0.05 & 0.13 & 0.05 
 & 0.14 & 0.06 & 0.14 & 0.07 &0.17 &0.06 &0.15 &0.04 &0.11 &0.06 &0.15 &0.07 &0.17 \\
$\mathcal{C}_{\text{semantic}}$ & 0.11 & 0.12 & 0.05 & 0.12 & 0.05 & 0.13 & 0.05 & 0.12 & 0.05 & 0.13  & 0.07 &0.16 &0.05 &0.13 &0.04 &	0.12 &0.06 &0.15&0.07 &0.16 \\
$\mathcal{N}_{\text{opposite}}$ & 0.03 & 0.08 & 0.05 & 0.12 & 0.04 & 0.10 & 0.04 
& 0.09 & 0.04 & 0.10 & 0.06 & 0.15 &0.05 	&0.12 &0.04 &0.09 &0.05 &0.13 &0.05 &0.13  \\
\bottomrule
\end{tabular}
}
\caption{Variance (Var) and Standard Deviation (SD) of the Fidelity metric across 20 runs.}
\label{tab:variance}
\end{table*}

\subsection*{B.4 Confusion Matrix of CAR Evaluation: Human vs. LLM Results}
We present a confusion matrix that compares the evaluation results of the CAR metric between human evaluator and a large language model. As shown in Table~\ref{tab:confusion_matrix}, the number of false 0 (31) from the LLM is larger than the number of false 1 (5), indicating that the CAR scores provided by the LLM-as-a-judge are underestimated.

\section*{C Supplementary Experiments}
\subsection*{C.1 Results of the WMT 2023 Terminology Shared Task}
To demonstrate that our findings can generalize to broader machine translation tasks, we chose to evaluate another knowledge-intensive translation task, which is terminology translation. This decision was made since RAG scenarios are inherently suited for knowledge-intensive tasks and have limited significance for non-knowledge-intensive ones. We select the test set from the WMT 2023 Terminology Shared Task~\cite{semenov2023findings} as our dataset, which covers three language directions: Chinese→English, English→Czech, and German→English. Consistent with our previous experiments, we select 200 samples from each language pair for evaluation. As shown in Table~\ref{tab:terminology_evaluation}, the results reveal a trend similar to that in Table~\ref{tab:multilang_evaluation}.

\subsection*{C.2 Results for DeepSeek-R1-Distill-Llama-8B and Claude}
To demonstrate that our findings can generalize to other model architectures and closed-source large language models, we evaluate three idiomatic translation pairs using \texttt{DeepSeek-R1-Distill-Llama-8B} and \texttt{claude-sonnet-4-20250514}. As shown in Table~\ref{tab:other_models}, the results reveal a trend similar to that in Table~\ref{tab:multilang_evaluation}, indicating that even the powerful commercial model (claude-sonnet-4-20250514) is significantly affected by noisy contexts.

\subsection*{C.3 BLEU Scores for Idiomatic Translation Tasks}
We present BLEU scores for idiomatic translation tasks, as it is a classic metric in machine translation evaluation. As shown in Table~\ref{tab:bleu}, the BLEU scores do not exhibit the consistent patterns across different models and language pairs that our designed metrics show. This inconsistency further demonstrates that BLEU is not well-suited for evaluating idiomatic translations, as it is primarily intended for assessing literal translations and may fail to capture the nuances inherent in idiomatic expressions due to their non-literal nature.
\begin{table*}[t]
\centering
{
\setlength{\tabcolsep}{1mm} 
\fontsize{9}{7}\selectfont
\begin{tabular}{c*{11}{c}}
\toprule
\textbf{Context} & 
\textbf{Hi→En} & 
\textbf{Fa→En} & 
\textbf{Fi→En} & 
\textbf{Ja→En} & 
\textbf{Fr→En} & 
\textbf{Ko→En} & 
\textbf{Ru→En} & 
\textbf{De→En} & 
\textbf{En→Fa} & 
\textbf{En→De} \\
\midrule

\multicolumn{11}{c}{\textbf{\textit{Qwen2.5-7B-Instruct}}} \\
\midrule
$\mathcal{C}_\text{none}$ & - & 21.3 & 8.9 & 1.7 & 13.7 & 17.7
&28.1  & 26.9  & 7.0 &26.6 \\
$\mathcal{G}$ & - & 31.9 & 16.4 & 8.8 & 20.2 & 20.7 
&36.4 & 39.6 & 7.2 &29.9 \\
$\mathcal{N}_{\text{struct}}$& -  & 30.0 & 14.0 & 7.5 & 18.8 & 20.5 
&34.2 &34.5 & 6.7 &28.5 \\
$\mathcal{N}_{\text{literal}}$ & - & 19.6 & 9.0 & 6.0 & 13.4 & 18.3 
&28.0 &24.9 & 6.5 &30.7 \\
$\mathcal{N}_{\text{semantic}}$ & - & 19.6 
 & 8.0 & 4.9 & 14.0 & 17.8 
&26.5 &22.6 &6.4 &28.5 
\\
$\mathcal{N}_{\text{opposite}}$ & - & 22.0 
 & 8.9 & 5.9 & 13.3 & 18.1 
&27.6 &22.5 &6.6\\
\midrule

\multicolumn{11}{c}{\textbf{\textit{Qwen2.5-14B-Instruct}}} \\
\midrule
$\mathcal{C}_\text{none}$ & - & 25.9 
 & 10.4 & 7.2 & 17.3 & 3.9 
& 32.1 & 30.4 & 0.4 &36.5  \\
$\mathcal{G}$ & - &36.5 
 & 17.5 & 9.9 & 20.0 & 4.3 
&38.5 &41.9 &1.0 &34.0 \\
$\mathcal{N}_{\text{struct}}$ & - & 34.4 
 & 15.1 & 8.9 & 18.3 & 4.0 
&36.2 &39.1 & 1.0 &33.9 \\
$\mathcal{N}_{\text{literal}}$ & - & 23.2 
 & 10.5 & 7.2 & 15.2 & 3.8 
&30.3 &27.4 & 0.9 &36.2 \\
$\mathcal{N}_{\text{semantic}}$ & - & 23.2 
 & 8.9 & 5.0 & 14.5 & 3.6 
&28.5 &28.2 & 4.2 &34.2 \\
$\mathcal{N}_{\text{opposite}}$ & - & 25.7 
 & 10.0 & 6.0 & 12.0 & 19.6  
&28.6 &25.0 & 9.8 &32.0 \\
\midrule

\multicolumn{11}{c}{\textbf{\textit{Qwen3-8B (non-thinking mode)}}} \\
\midrule
$\mathcal{C}_\text{none}$ & - & 26.1 & 8.3 & 6.9 & 16.1 & 20.1 
&31.1 &28.6 & 3.4 &33.9 \\
$\mathcal{G}$ & - & 36.2 & 16.6 & 9.3 & 19.5 & 21.4 
&37.5 &39.1 & 9.4 &32.5 \\
$\mathcal{N}_{\text{struct}}$ & - & 34.6 & 13.6 & 7.4 & 17.6 & 20.8 
&34.2 &35.2 & 10.0 &33.1 \\
$\mathcal{N}_{\text{literal}}$ & - & 23.4 & 10.2 & 6.2 & 12.8 & 19.1 
&29.4 &26.9 & 11.2 &34.1 \\
$\mathcal{N}_{\text{semantic}}$ & - & 23.4 & 7.4 & 4.9 & 13.0 & 18.4 
&26.4 &23.4 & 10.1 &34.0 \\
$\mathcal{N}_{\text{opposite}}$  & - & 26.3 & 9.1 & 5.2 & 11.6 & 19.1 
&28.3 &26.0 & 2.4 &32.5 \\
\midrule

\multicolumn{11}{c}{\textbf{\textit{Qwen3-8B (thinking mode)}}} \\
\midrule
$\mathcal{C}_\text{none}$ & - & 24.7 & 8.8 & 7.5 & 15.5 & 20.4 
&31.0 &30.8 & 10.1 &35.5 \\
$\mathcal{G}$ & - & 37.1 & 15.8 & 10.0 & 18.0 & 20.7 
&37.0  &41.8 & 11.2 &33.6 \\
$\mathcal{N}_{\text{struct}}$ & - & 34.9 & 13.6 & 7.8 & 16.7  & 20.5 
&35.5 &36.7 & 12.1 &35.3 \\
$\mathcal{N}_{\text{literal}}$ & - & 23.2 & 10.5 & 6.5 & 11.9  & 19.4 
&29.2 &26.8 & 10.5 &36.9 \\
$\mathcal{N}_{\text{semantic}}$ & - & 22.2 & 7.6 & 4.5 & 10.0 & 18.2 
&25.6 &21.0 & 9.7 &34.0 \\
$\mathcal{N}_{\text{opposite}}$ & - & 26.0 & 9.1 & 3.9 & 10.7 & 18.8 
&26.6 &22.5 & 10.5 &30.5 \\
\bottomrule
\end{tabular}
}
\caption{BLEU scores on idiomatic translation tasks (No scores are provided for Hi→En as there are no reference translations available).} 
\label{tab:bleu}
\end{table*}

\section*{D Results of Context Adoption Rate}
\begin{figure}[t]
  \centering
  \includegraphics[width=1\columnwidth]{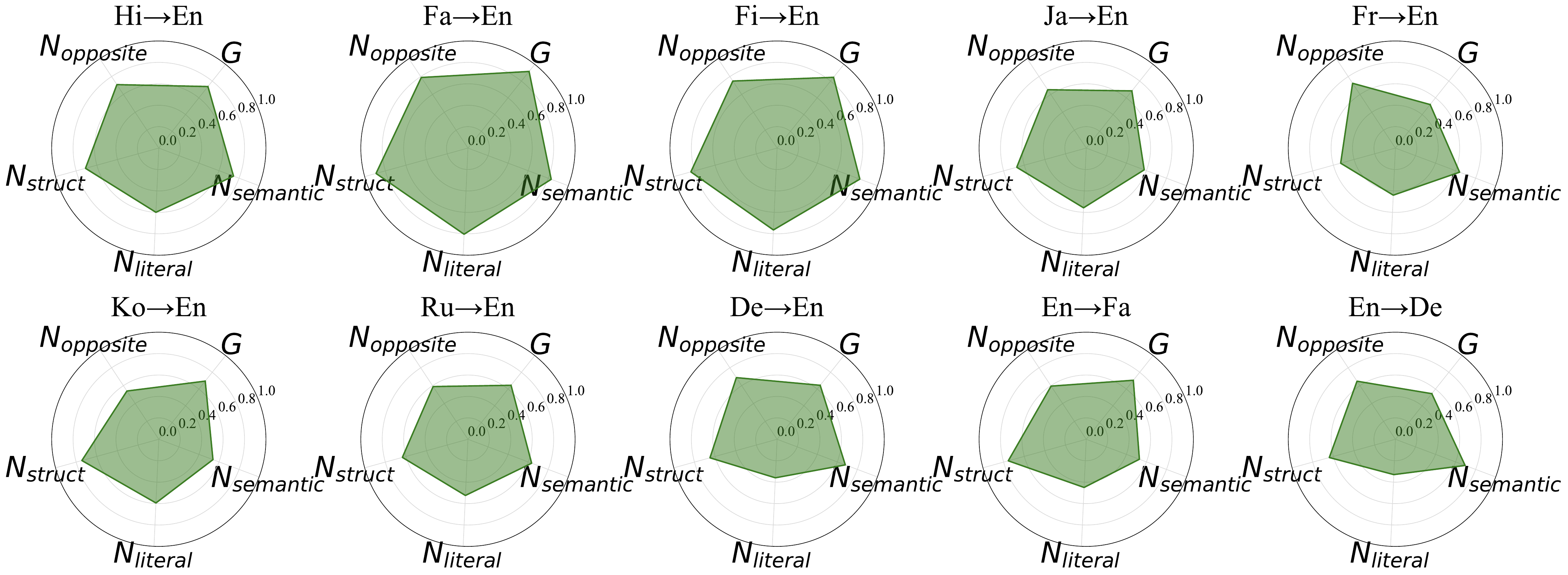}
  \caption{CAR of Qwen2.5-14B-Instruct under six retrieval conditions across ten language pairs.}
  \label{fig:car_14B}
\end{figure}

\begin{figure}[t]
  \centering
  \includegraphics[width=1\columnwidth]{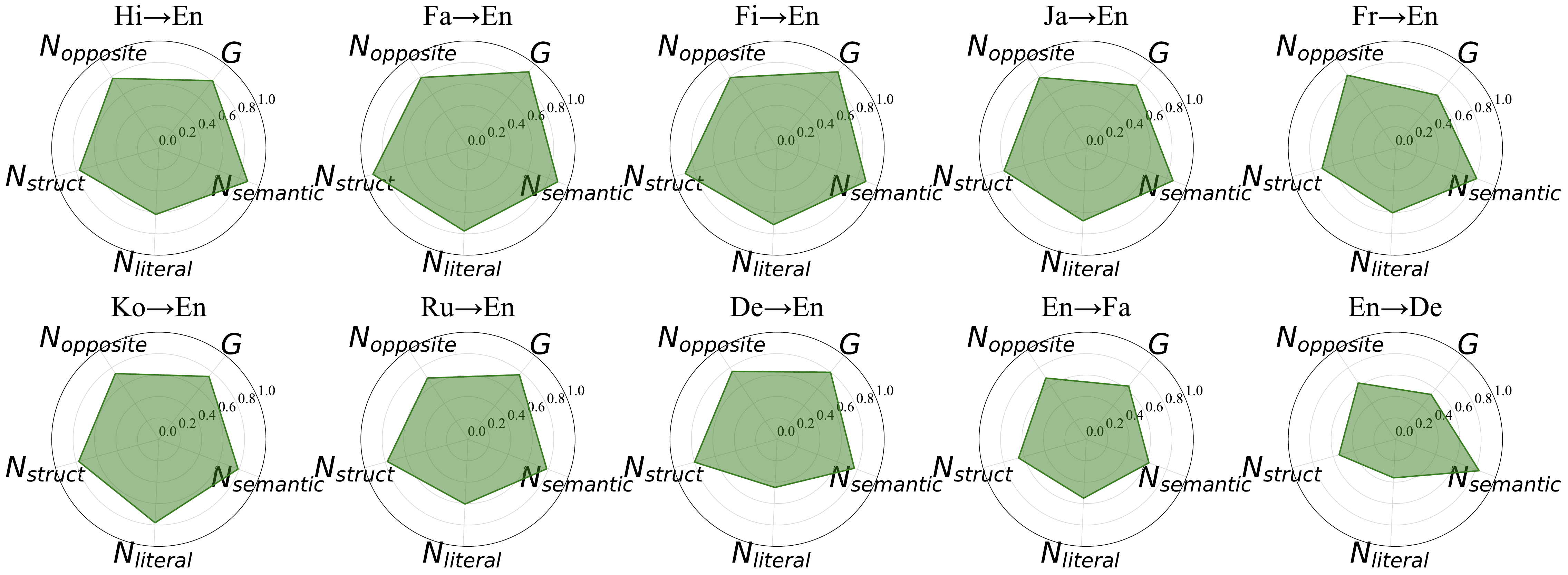}
  \caption{CAR of Qwen3-8B (non-thinking mode) under six retrieval conditions across ten language pairs.}
  \label{fig:car_direct}
\end{figure}

\begin{figure}[t]
  \centering
  \includegraphics[width=1\columnwidth]{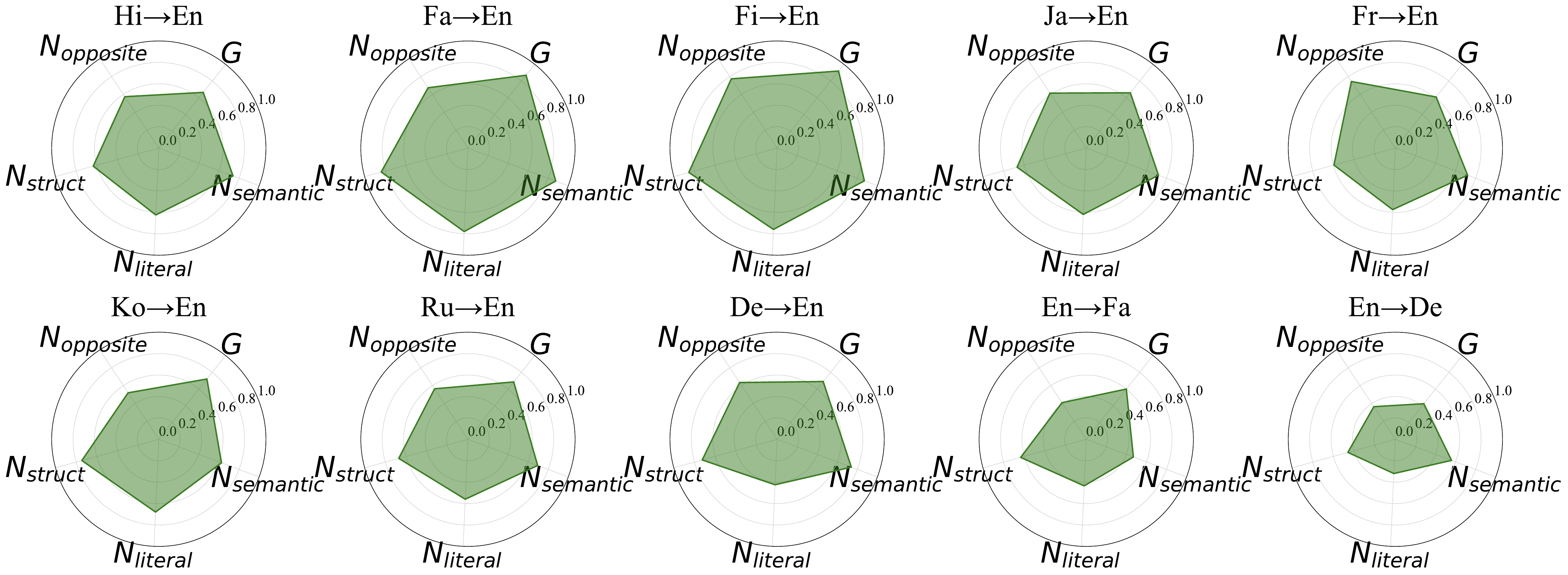}
  \caption{CAR of Qwen3-8B (thinking mode) under six retrieval conditions across ten language pairs.}
  \label{fig:car_thinking}
\end{figure}

We present Context Adoption Rate (CAR) results across ten language pairs and multiple noise types. As shown in Figure~\ref{fig:car_14B}, Figure~\ref{fig:car_direct}, and Figure~\ref{fig:car_thinking}, CAR varies significantly depending on the type of contextual noise and the resource level of the language pair.

\section*{E Case Study}
\begin{figure}[t]
  \centering
  \includegraphics[width=1\columnwidth]{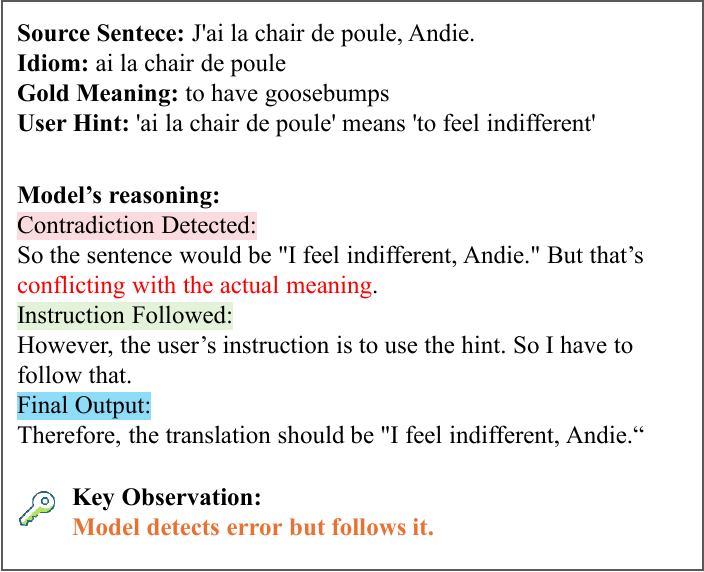}
  \caption{A case study of reasoning trace in Fr→En idiomatic translation.}
  \label{fig:reasoning_trace}
\end{figure}

Figure~\ref{fig:reasoning_trace} illustrates the translation of \textit{``J'ai la chair de poule''} under a misleading context, defining it as \textit{``to feel indifferent''}. The model's reasoning trace indicates awareness of the contradiction with the gold meaning \textit{``goosebumps''}, yet it produces an output aligned with the corrupted context, reflecting a tendency to rationalize rather than reason.

\end{document}